\documentclass{article}

\usepackage{microtype}
\usepackage{graphicx}
\usepackage{subcaption}
\usepackage{booktabs} 

\usepackage{hyperref}
\usepackage{adjustbox}

\usepackage[accepted]{icml2026}

\usepackage{amsmath}
\usepackage{amssymb}
\usepackage{mathtools}
\usepackage{amsthm}
\usepackage{times}
\usepackage{latexsym}
\usepackage[T1]{fontenc}
\usepackage[utf8]{inputenc}
\usepackage{microtype}
\usepackage{inconsolata}
\usepackage{booktabs}
\usepackage{amsmath}
\usepackage{hyperref}
\usepackage{multirow}
\usepackage{booktabs}
\usepackage[table]{xcolor}
\usepackage{tabularx}

\usepackage{graphicx}

\newcommand\sect[1]{\S\ref{#1}}

\usepackage[capitalize,noabbrev]{cleveref}

\theoremstyle{plain}

\theoremstyle{definition}

\theoremstyle{remark}

\usepackage[textsize=tiny]{todonotes}

\icmltitlerunning{Evaluating Pluralism in LLMs through Latent Perspectives}

\begin{document}

\twocolumn[
  \icmltitle{Evaluating Pluralism in LLMs through Latent Perspectives}

  \icmlsetsymbol{equal}{*}

  \begin{icmlauthorlist}
    \icmlauthor{Laura Majer}{yyy}
    \icmlauthor{Jan Šnajder}{yyy}
    \icmlauthor{Martin Tutek}{yyy}
  \end{icmlauthorlist}

  \icmlaffiliation{yyy}{TakeLab, University of Zagreb, Croatia}

  \icmlcorrespondingauthor{Laura Majer}{laura.majer@fer.hr}

  \icmlkeywords{Machine Learning, ICML}

  \vskip 0.3in
]

\printAffiliationsAndNotice{}  

\begin{abstract}
   The growing need to represent diverse perspectives has increased interest in pluralistic LLM generation. Although difficult to operationalize, identifying \textit{perspectives} expressed in text would provide clear guidance on pluralistic alignment and more clearly articulate the pluralistic gap in LLM generation. While models have been shown to reduce the diversity of training data and generate homogeneously, this has been demonstrated primarily on multiple-choice questionnaires or using high-level characteristics of free-form text. In this paper, we introduce and implement a domain-agnostic multi-layered framework for unsupervised extraction of perspectives suitable for identifying the pluralistic gap in LLM-generated text. We evaluate our framework on book reviews, a highly opinionated dataset representing diverse perspectives, and compare various prompts and models. Our results show that while some models and prompting techniques come close to covering a broad spectrum of perspectives, rarer perspectives remain disproportionately underrepresented, resulting in distributions that diverge from human text.

\end{abstract}

\section{Introduction}
As large language models scale, their competence across a broad range of tasks such as coding, math, and complex reasoning also improves \citep{jimenez2024swebench,phan2025humanity}.
However, performance on tasks with objectively verifiable outputs does not necessarily indicate alignment to the inherent diversity of human perspectives in the myriad of subjective tasks where ``correctness'' is defined by specific cultural, social, or individual frameworks rather than universal facts \citep{frenda2025perspectivist, fleisig-etal-2024-perspectivist}.
To provide nuanced and balanced viewpoints on open-ended subjective tasks, language models should undergo \textit{pluralistic alignment} \citep{roadmap2024sorensen}, that is, they should be optimized to adequately represent the diversity of perspectives of a target population in their generated responses. Concerningly, current empirical evidence suggests that frontier models fall short of this ideal: LMs exhibit a lower linguistic variety \citep{russo2025pluralisticmoralgapunderstanding} and a high homogeneity in their generated texts -- a phenomenon dubbed the \textit{generative monoculture} \citep{wu2025generative}.
In general, models produce both inter- and intra-homogeneous outputs, generating repetitive text that is also highly similar across model families \citep{jiang2025artificial}. 

\begin{figure}
    \centering
    \includegraphics[width=1\linewidth]{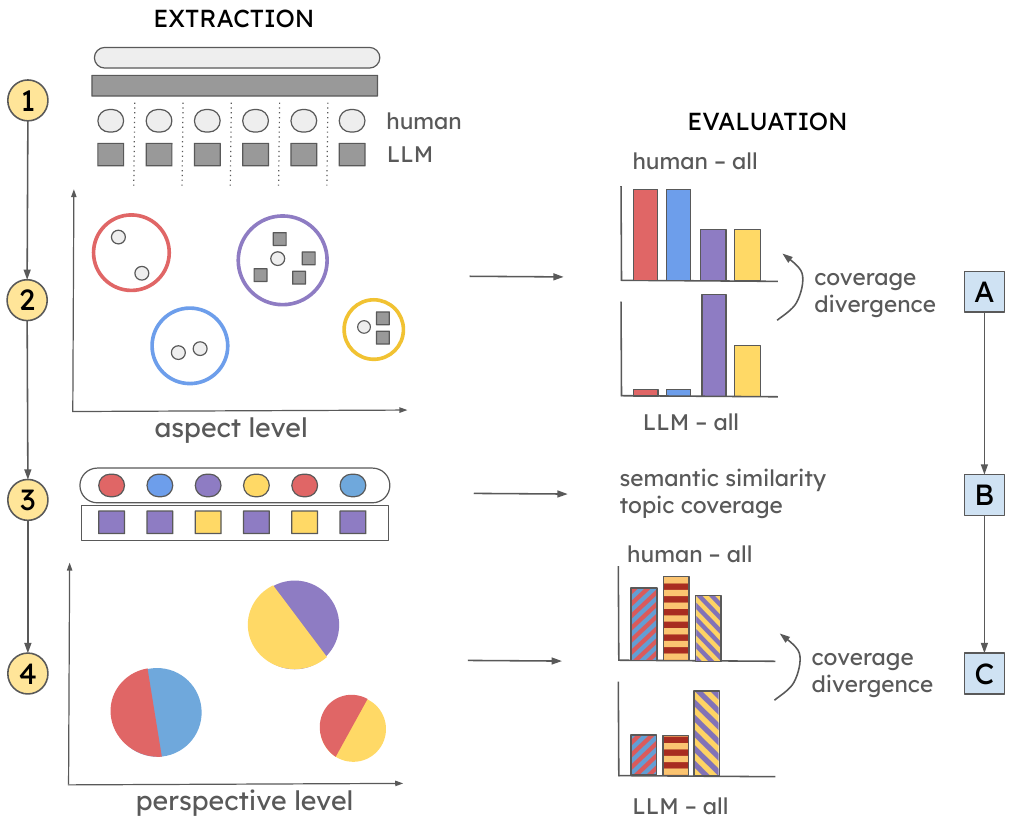}
    \caption{The proposed pluralistic evaluation framework, used for extracting perspectives and evaluating their diversity in human and LLM-generated data. We first identify aspects from text (1), cluster them (2), producing perspective representations (3), which we cluster again to identify collective perspectives (4). Across levels, we evaluate aspect level coverage (A), features of the perspective representation (B), and perspective coverage (C).}
    \label{fig:infograph}
\end{figure}

One way to assess the degree of pluralism represented by a model is to analyse the diversity of \emph {perspectives} represented in the outputs it generates. Perspective, in the most abstract sense defined as ``a particular way of considering something'' \citep{cambridge_perspective}, is embedded in every communication act \citep{basile2022perspective}. In NLP, perspectives are generally used as an umbrella term for subjective language and are considered a facet of pluralistic alignment \citep{roadmap2024sorensen}.
Although perspective is difficult to operationalize, it is the \textit{latent driver} behind overt manifestations such as sentiment, opinions, and claims, and it provides a powerful framework for evaluating model pluralism by unifying these otherwise disparate constructs into a holistic lens.

The ability of LLMs to model human perspectives was previously evaluated primarily using MCQ surveys \cite{santurkar2023opinions} and Likert scales \cite{meister2025benchmarking}. Where free-form text has been considered, homogeneity has been measured using aggregated, high-level features \cite{wu2025generative}, with little attention to the range of perspectives expressed in human text. By reducing perspectives to aggregated labels, prior approaches fail to model the very complexity they seek to measure, ultimately masking the extent of the \textit{pluralistic gap} in model generation.

To address these limitations, we introduce a domain-agnostic framework that formalizes \emph{latent perspective} as a composite of aspects extracted from text, offering a fine-grained alternative to surface-level homogeneity metrics. By using aspects as building blocks to characterize the collective perspectives expressed in free-form text, our method enables a high-resolution comparison of how perspective distributions vary between any diverse text collections, but is particularly suitable for human and LLM-generated content. A visualization of our framework is shown in \Cref{fig:infograph}. We test the applicability of our framework on a corpus that is diverse, opinionated, and with clear topics -- a book review dataset sourced from the Goodreads platform \citep{goodreads2018recommender}. We evaluate pluralism across open-source (\texttt{Llama 3 8B}, \texttt{OLMo2 1B \& 8B}) and proprietary (\texttt{GPT 4.1} and \texttt{Gemini 2.5}) models. To elicit higher coverage of perspectives in LLM outputs, we use high temperature as well as persona prompting \citep{ge2025personas}, a technique applied heavily in population simulation. We apply our framework to the LLM-generated reviews and assess their pluralism using various quantitative distributional measures, including divergence and coverage, as well as a qualitative analysis to identify topics underrepresented in LLM text.

Our framework allows for a unified view of both the plurality of perspectives present in long-form LLM outputs, and the comparison with human text. Our results confirm previous findings regarding the surface-level homogeneity of LLM outputs and also demonstrate that when evaluating pluralism in the context of perspective, models vary significantly in both baseline performance and their sensitivity to prompting. We find that, while some models and configurations might come close to encompassing a spectrum of diverse aspects (Overton pluralism), rarer aspects remain disproportionately underrepresented (distributional pluralism), leading to a homogeneity in majority perspectives. Our contributions can be summarized as follows: \textbf{(i)} we propose a multi-leveled framework for comparing latent perspectives expressed in two sources of text, suitable for analysing model pluralism, \textbf{(ii)} we implement the framework on the opinionated dataset of book reviews across different LLMs and prompting methods designed to elicit diverse outputs, and \textbf{(iii)} we analyse perspective gaps at various levels of abstraction and differences across models and prompting techniques.  
\section{Related Work}
\label{sec:RW}

The tendency of LLMs to flatten the diversity of human perspectives by converging toward a more uniform distribution of features and semantics in their outputs has been studied under several names, including \textit{artificial hivemind} \citep{jiang2025artificial}, \textit{generative monoculture} \citep{wu2025generative}, and \textit{distributional gap} \citep{peeperkorn2025gap}, and reframed constructively as the effort for higher model \textit{pluralism}. \citet{roadmap2024sorensen} propose three pluralistic alignment modes: \textit{Overton}, representing the set of possible answers, \textit{distributional}, representing the distribution of possible answers, and \textit{steerable}, representing output similar to a chosen group to steer towards. Evaluation of these modes in prior work is predominantly discrete: opinion alignment is operationalised through MCQ choices or Likert ratings \citep{santurkar2023opinions, durmus2024towards, meister2025benchmarking}, or reduced to classification accuracy over attribute profiles and preference rankings \citep{adams2025steerable, chen2025spica}.
These approaches share a common limitation by reporting aggregate statistics over a population, discarding fine-grained information indicating how opinions are framed and which aspects are pivotal.

Fewer works study output homogenisation directly on the features of free-form text, as opposed to the projected labels. \citet{wu2025generative} demonstrate generative monoculture in book reviews by measuring diversity through extracted attributes (binary sentiment and coarse topic labels), while \citet{peeperkorn2025gap} measure the diversity gap in narrative generation via aggregate scalar scores such as the Vendi Score. While both argue that LLM-generated text exhibits a more narrow distribution of features compared to human text on the matching tasks and topics, their metrics collapse the distributional structure of generated text into a single diversity score, making it difficult to identify which perspectives are systematically absent. A deeper, more structured evaluation is necessary, one that takes into account the multifaceted properties of perspective. 

The limited evaluation on free-form text largely reflects the difficulty of working with it. Even within perspectivism \cite{frenda2025perspectivist} -- a paradigm that emphasizes preserving diverse perspectives across NLP tasks -- perspective is typically operationalized as task-specific labels assigned to instances. These labels can then support various modeling and analysis approaches, such as creating annotator profiles using clustering \cite{vitsakis-etal-2024-voices}, but perspective is still effectively anchored to a label space, leaving the nuances of free-form expression largely unaddressed. Clustering has been used as a method of identifying common elements in various subjective tasks, both in argumentation \citep{reimers-etal-2019-classification} and framing \citep{ajjour-etal-2019-modeling}, though it has not been used for measuring and comparing the the homogeneity of LLMs, a gap which we fill with this paper.
\section{A Framework for Perspective Analysis}
\label{sec:framework}

\subsection{Motivation}
\label{sec:motivation}

Perspectives in text are expressed through inherently subjective language, influenced by individual interpretation, emphasis, and context. A prerequisite for automatic extraction of perspective is operationalizing it, which is difficult because precise definitions and appropriate levels of granularity are context-sensitive.
This results in two central challenges: (1) how to construct a meaningful representation of diverse perspectives on a given topic and (2) how to compare the distribution of perspectives from different data sources, such as human and LLM-generated text.

To address these challenges, we propose a two-level framework for identifying perspectives identified in a collection of texts and assessing their diversity. The first level operates at the level of \textit{aspects}, where the representations of individual discourse units from texts are isolated and grouped to capture recurring semantic patterns. The second level operates at the \textit{perspective level}, using coarse labels of aspect groups produced at the previous level to characterize the perspective, both underlying and verbalized in the text. Our framework is grounded in the idea of collective perspectives, identifying recurring patterns expressed across individuals that represent both majority and minority viewpoints, without imposing constraints on their form or content.

The proposed framework enables a coarse-grained analysis of perspectives from free-form text, as well as a fine-grained comparison of the underlying aspects at the first level. A concrete implementation of the framework requires operationalizing its core components: the first level of aspect segmentation and identification of collective aspects, the second level where perspectives are constructed from aspects and grouped, and defining metrics that compare perspective diversity from different sources.
In the following section, we provide the formal definition of the framework (\sect{sec:definition}), followed by a concrete implementation (\sect{sec:implementation}), which we then apply to a book corpus (\sect{sec:results}) and finally rigorously verify the validity of its components (\sect{sec:analysis}).

\subsection{Definition}
\label{sec:definition}

Let $D = \{d_1, d_2, \ldots, d_n\}$ be a dataset of texts sharing a common topic, such as reviews of the same book. 

\paragraph{Aspects.}
Each text $d_i$ can be decomposed into a sequence of $m_i$ aspect instances: 
\begin{equation}
d_i = [a^i_1, a^i_2, \ldots, a^i_{m_i}]
\end{equation}
where each $a^i_j$ represents the aspect expressed in unit $j$ of text $i$. Common aspects should then be grouped across $D$, yielding a set of $k$ aspect clusters $\mathcal{C} = \{C_1, \ldots, C_k\}$. Each aspect instance $a^i_j$ is assigned to exactly one cluster $C_l \in \mathcal{C}$, or designated as an outlier and excluded:
\begin{equation}
a^i_j \mapsto \begin{cases} C_l & \text{if } a^i_j \text{ belongs to a recognised cluster} \\ \varnothing & \text{if } a^i_j \text{ is an outlier} \end{cases}
\end{equation}

\paragraph{Perspective.}
We define the perspective of a text $d_i$ as the distribution over clusters induced by its non-outlier aspect instances. Let $n^i_l$ denote the number of aspect instances in $d_i$ assigned to cluster $C_l$, and let $N_i = \sum_{l=1}^{k} n^i_l$ be the total number of non-outlier aspect instances. We define the perspective of $d_i$ as the vector:
\begin{equation}
\mathbf{p}_i = \left( \frac{n^i_1}{N_i},\ \frac{n^i_2}{N_i},\ \ldots,\ \frac{n^i_k}{N_i} \right) \in \Delta^{k-1}
\end{equation}
where $\Delta^{k-1}$ denotes the $(k-1)$-dimensional probability simplex.

\subsection{An Implementation of the Framework}
\label{sec:implementation}

\subsubsection{Aspect level}

In our work, we choose a sentence as the target discourse unit representing an aspect, and use sentence embeddings as its semantic representation within the framework. We do not construct a structured representation of aspects, which keeps the approach context-independent. We provide a more structured analysis (\sect{sec:topic_evaluation}) and validate the quality of this representation (\sect{sec:cluster_coherence}) in later sections.
To estimate the distribution of aspects present in book reviews, we (1) segment the original reviews into sentences, then (2) embed each sentence using a sentence encoder model, and finally (3) group the embeddings into aspect clusters.

\paragraph{Implementation.} We use spaCy to segment free-form text into sentences.\footnote{\url{https://spacy.io/models/en}} We opt for the F2LLM 0.6B model \citep{zhang2025f2llmproc} as the sentence embedder, as it currently achieves the best clustering performance among sub-1B parameter English models on the MTEB leaderboard.\footnote{\url{https://huggingface.co/spaces/mteb/leaderboard}}
To cluster the aspects encoded within sentences, we build on the BERTopic \citep{grootendorst2022bertopic} library, and choose HDBSCAN \citep{HDBSCAN} as the aspect-level clustering algorithm, as unlike centroid-based methods such as k-means, it does not require the number of clusters to be specified a priori. This property is crucial in our setting, as the number of collective aspect clusters in a group of reviews is not known beforehand, and predefining it would impose unnecessary constraints. Prior to clustering, we reduce the dimensionality of sentence embeddings to $5$ using UMAP \citep{McInnes2018}.\footnote{We ran experiments covering higher dimensionality with comparable or worse results.}

We select 1500 human reviews per book for our analysis, and split them into a fixed base ($R_\mathit{base}$) and evaluation ($R_{eval}$) set, with $|R_\mathit{base}| = 1000$ and $|R_\mathit{eval}| = 500$.
We compute the clusters on the base set, and use the held-out evaluation set for comparison with the same number of LLM-generated reviews. We opt for this split to ensure a larger set of reviews that serves both as a stable base for cluster computation and as a more accurate approximation of the full aspect distribution, while the smaller number of held-out reviews allows for an apples-to-apples comparison with LLM-generated reviews. The sentencized and embedded reviews from $R_\mathit{base}$ are used to fit a clustering model for each book ($b$), resulting in $K_b$ clusters per book.
Sentences from $R_\mathit{eval}$ and LLM-generated reviews are then assigned aspect cluster labels using the fit clustering models.

\paragraph{Evaluation.}

We use two quantitative metrics to evaluate aspect-level differences between human and LLM-generated texts.
If LLMs are properly pluralistically aligned, the cluster assignment compared to human texts should be (1) complete, covering all clusters to reflect \emph{Overton pluralism}, and (2) proportionate, maintaining the same ratios to reflect \emph{distributional pluralism}. We operationalize these using cluster coverage percentage and Jensen-Shannon Divergence (JSD), respectively, measured between the cluster distributions at the book level, then averaged across all books. 

Apart from these contrastive measures, we also compare individual population statistics, which measure aspect diversity within reviews originating from a single source. As these statistics, we opt for average semantic similarity, measuring sentence homogeneity, and aspect cluster entropy. 

\subsubsection{Perspective level}
To aggregate the individual aspects to the perspective level, we encode each review as a set of aspects based on the cluster assignments of its constituent sentences. Since HDBSCAN is a probabilistic model, its output can be interpreted either as a distribution over $K_b$ aspect clusters or as the most probable cluster. We opt to use the most probable cluster as the label, constructing the perspective representation as the count of aspects in that cluster. The individual representations are then clustered to recognise collective perspectives.

\paragraph{Implementation.}
After constructing the perspective vectors in the same human ($R_\mathit{base}$, $R_\mathit{eval}$) and LLM-generated sets, we implement another level of clustering models to determine the collective perspectives. In contrast to the aspect level, perspective vectors are sparser, making the previous density-based HDBSCAN model no longer effective. Therefore, we utilize other clustering models, including k-means and community detection. 

\paragraph{Evaluation.}
Akin to the aspect level, we evaluate the distribution of perspectives using total cluster coverage and the Jensen-Shannon Divergence (JSD) to explicitly compare Overton and distributional pluralism. We complement this by cosine similarity of the perspective representation vectors to assess population diversity.  

\section{Dataset and Models}
\label{sec:methodology}

\subsection{Dataset}
We utilize the Goodreads dataset \citep{goodreads2018recommender} for all experiments. The dataset consists of book reviews and corresponding metadata across various genres. 
We chose this dataset because it contains opinionated texts not explicitly related to political stances. Furthermore, the 2017 data cut-off nullifies the risk of LLM-generated text.

For our analysis, we select from English-language books that contain more than 1500 reviews. To ensure a representative sample, we sample 20 books from each genre, including five from each of the following categories: highest average score, lowest average score, highest score deviation, and highest number of reviews. The represented genres are \textit{Mystery, Thriller and Crime}, \textit{Young Adult}, \textit{History and Biography}, \textit{Fantasy and Paranormal}, and \textit{Romance}.

From these, we select a preliminary \textit{analysis subset} of 10 books, sampled across genres, for initial testing and analyses. 
We filter out reviews not written in English or those shorter than 20 characters, using \textit{langdetect} for language detection.\footnote{\url{https://pypi.org/project/langdetect/}} 

\subsection{Models and prompting configurations}
We run our experiments across various closed- and open-source LLMs. For closed-source models, we select \texttt{gpt-4.1-mini-2025-04-14} \citep {achiam2023gpt} and \texttt{gemini-flash-2.5-mini} \citep{comanici2025gemini} (hereafter abbreviated as \texttt{GPT 4.1} and \texttt{Gemini 2.5}). For open-source models, we select \texttt{Llama-3.1-8B} \citep{grattafiori2024llama} as well as \texttt{OLMo-2-1B} and \texttt{OLMo-2-8B} \citep{olmo20252olmo2furious}. The availability of pre-training corpora for OLMo models allows us to verify whether they were exposed to the review texts. 

We measure and compare the diversity of perspectives in book reviews generated by humans and LLMs under different prompting setups. We generate 300 reviews for the 10-book subset and 100 reviews for the full 100-book set.
We utilize three different prompting setups:

\paragraph{Baseline.} \textit{Baseline} prompts contain the vanilla instruction to write a book review, as well as rate it. We use a temperature $T=0.7$ to obtain variety across samples.

\paragraph{High temperature.} Higher temperature values are often used in creative tasks \citep{wu2025generative}.
To evaluate the influence of higher temperature on review diversity, we use the baseline prompt and evaluate it at $T=1.5$.

\paragraph{Persona prompts.} Socio-demographic prompting and persona prompting are promising ways to increase the diversity of LLM output. 
We select a fixed batch of $300$ personas from \citet{ge2025personas} and use each persona to generate a single review.

We report prompts used across all setups in Appendix \ref{sec:prompts}.

\section{Results}
\label{sec:results}

In this section, we first present aggregate measures (\sect{sec:aggregate}) and then fine-grained aspect- and perspective-level comparisons from our framework (\sect{sec:framework_results}).

\subsection{High-level measures}
\label{sec:aggregate}

We first perform a set of analyses focusing on high-level, aggregated measures to assess consistency with prior findings and to confirm differences in superficial features between the baseline and other setups.
We report results on the analysis subset of 10 books, comparing 300 model-generated reviews with 300 held-out human reviews from $R_\mathit{eval}$.

\paragraph{Semantic similarity.} To measure the novelty of generated reviews across runs, we process reviews for each book in sequential batches of 10. For each new batch, we compute the cosine similarity between each sentence embedding in the incoming batch and all sentence embeddings from previous batches, recording the maximum value. We then average this score across books to obtain a mean-max similarity score. A high mean-max similarity indicates that new reviews are progressively more repetitive, while a low value suggests they introduce novel content.

\begin{figure}
    \centering
    \includegraphics[width=\linewidth]{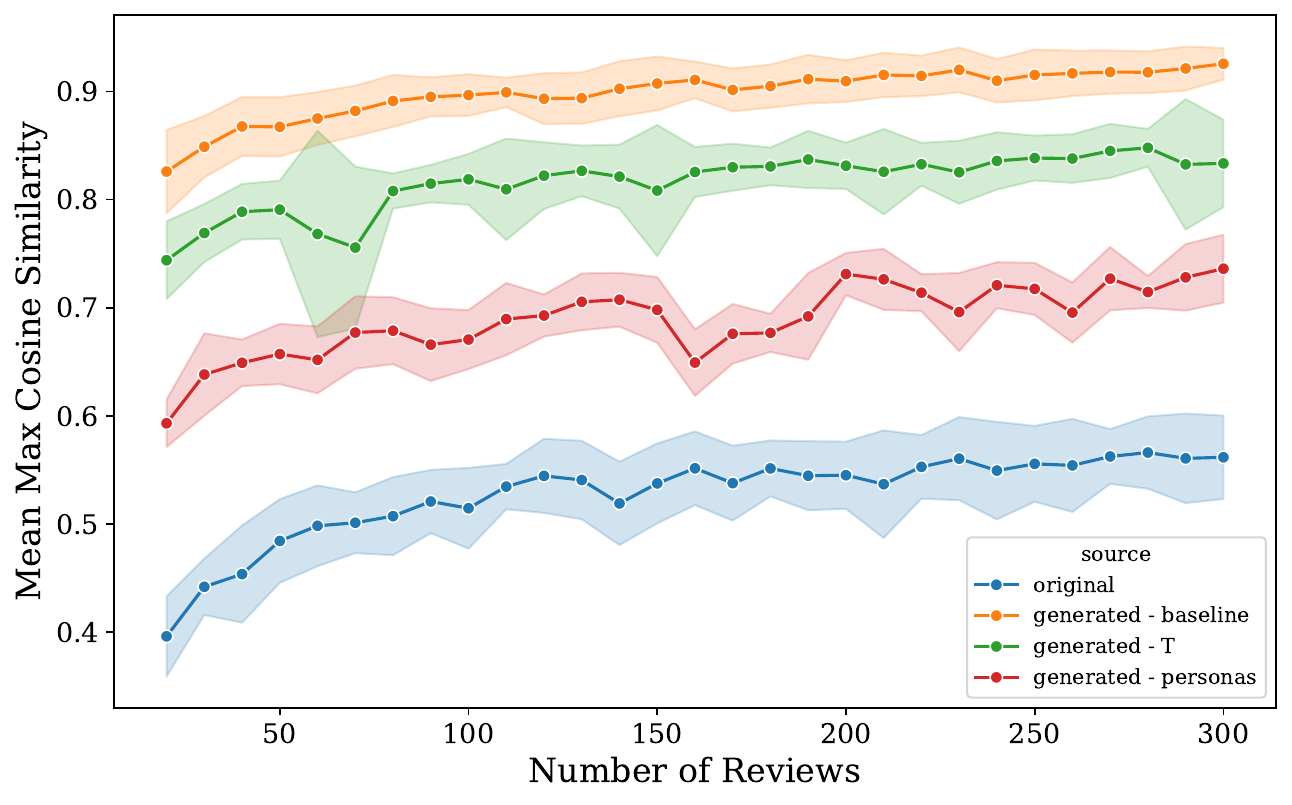}
    \caption{Mean maximum cosine similarity between embeddings for original reviews and across generation configurations for \texttt{GPT 4.1}. Results for other models are presented in Appendix \ref{sec:semantic_similarity}.}
    \label{fig:similarity}
\end{figure}

Evaluating the original and generated reviews separately allows us to compare how quickly each source saturates. \Cref{fig:similarity} shows the mean-max similarity curves for human reviews and those generated by \texttt{GPT 4.1}. 
The results point to two major findings: (1) diversity saturates at around 100 reviews, leading us to opt for that sample size in further experiments, and (2) the semantic similarity of human reviews is lower than that of generated reviews, with \textit{persona} prompting having a positive influence on review diversity. We report the results for other models in Appendix \ref{sec:semantic_similarity}.

\paragraph{Sentiment and rating.}
The review dataset is highly subjective, where attitudes toward a book are conveyed both implicitly through sentiment and explicitly through assigned ratings ranging from 1 to 5. \citet{wu2025generative} show LLM-generated reviews are more positive in sentiment than original reviews, but do not compare the assigned rating. We estimate sentiment of book reviews using \texttt{distilbert-base-uncased-finetuned-sst-2-english} \cite{sanh2019distilbert}. The sentiment range is $[-1,1]$, with $-1$ denoting negative and $+1$ positive sentiment, respectively.

We report the average sentiment and assigned ratings across books for original and generated reviews in \Cref{tab:combined_stats}. Results differ across models, with \texttt{Gemini 2.5} producing the lowest average rating and sentiment, and \texttt{GPT 4.1} the highest. Similar to our findings for semantic similarity, the baseline prompts are farthest from the original scores, scoring highly on both average sentiment (a trend most pronounced for \texttt{GPT 4.1}) and average assigned rating.
Higher-temperature sampling ($T=1.5$) yields mixed results, typically reducing the average rating and sentiment while increasing the corresponding standard deviation.
Persona prompting generates ratings closest to the original scores; notably, \texttt{Gemini 2.5} produces lower average ratings than the original reviews. 

\begin{table}[ht]
\caption{Mean$_{\pm\text{std}}$ book review rating and detected sentiment per model and generation configuration. 
}
\centering
\resizebox{\columnwidth}{!}{%
\begin{tabular}{llccccc}
\toprule
& & \text{GPT 4.1} & \text{Gemini 2.5} & \text{Llama3 8B} & \text{OLMo2 1B} & \text{OLMo2 7B} \\
\midrule
\multirow{4}{*}{\rotatebox[origin=c]{90}{\textbf{Rating}}}
& \text{base} & $4.07_{{\pm.57}}$ & $3.40_{{\pm.86}}$ & $4.11_{{\pm.35}}$ & $3.11_{{\pm.73}}$ & $4.16_{{\pm.41}}$ \\
& \text{temp} & $4.09_{{\pm.56}}$ & $3.46_{{\pm.88}}$ & $4.08_{{\pm.47}}$ & $3.15_{{\pm.99}}$ & $4.20_{{\pm.58}}$ \\
& \text{pers} & $3.48_{{\pm.70}}$ & $3.02_{{\pm.98}}$ & $3.58_{{\pm.69}}$ & $3.21_{{\pm.74}}$ & $3.64_{{\pm.76}}$ \\
\cmidrule(l){2-7}
& \cellcolor{gray!20}\text{original} & \multicolumn{5}{c}{\cellcolor{gray!20}$3.52_{{\pm1.30}}$} \\
\midrule
\midrule
\multirow{3}{*}{\rotatebox[origin=c]{90}{\textbf{Sentiment}}}
& \text{baseline} & $0.92_{{\pm.39}}$ & $0.45_{{\pm.87}}$ & $0.67_{{\pm.72}}$ & $0.34_{{\pm.92}}$ & $0.91_{{\pm.40}}$ \\
& \text{temp} & $0.89_{{\pm.44}}$ & $0.46_{{\pm.86}}$ & $0.67_{{\pm.72}}$ & $0.28_{{\pm.93}}$ & $0.84_{{\pm.51}}$ \\
& \text{personas} & $0.66_{{\pm.73}}$ & $0.26_{{\pm.92}}$ & $0.37_{{\pm.90}}$ & $0.41_{{\pm.89}}$ & $0.70_{{\pm.70}}$ \\
\cmidrule(l){2-7}
& \cellcolor{gray!20}\text{original} & \multicolumn{5}{c}{\cellcolor{gray!20}$0.03_{{\pm.95}}$} \\
\bottomrule
\end{tabular}}
\label{tab:combined_stats}
\end{table}

\subsection{Framework results}
\label{sec:framework_results}

\paragraph{Aspect level.}

To compare the topic distributions of human and LLM-generated reviews, we use two measures. First, we estimate alignment with Overton pluralism, reflecting the breadth of the generating corpus through \textit{cluster coverage}, which measures the percentage of human-identified clusters that appear at least once in the generated reviews.
Second, to estimate distributional pluralism -- whether LLM-generated aspects are discussed in proportions similar to those of human reviewers -- we compute the JSD, quantifying how much the generated corpus diverges from the human baseline in aspect distributions.

We report results across models and configurations in \Cref{tab:coverage_jsd}. Persona prompting consistently yields higher topic coverage and lower divergence from the human distribution than temperature-based sampling across all models and metrics. \texttt{Gemini-2.5} under persona prompting reaches the highest coverage ($98.0 \pm 2.6\%$) and lowest divergence ($\text{JSD} = 0.11 \pm 0.02$), suggesting it most closely approximates both Overton and distributional pluralism. Interestingly, \texttt{GPT-4} lags behind both \texttt{Gemini-2.5} and the smaller \texttt{Llama-3.1-8B} across all configurations, peaking at only $62.7 \pm 3.4\%$ coverage under persona prompting. A higher temperature ($T = 1.5$ vs.\ $T = 0.7$) yields modest but consistent gains in both metrics. Overall, we find that the choice of prompting strategy has a stronger effect on topical diversity than temperature scaling and that persona prompting can effectively narrow the pluralistic gap.

\begin{table}[ht]
    \centering
    \caption{Cluster coverage and JSD (mean$_{\pm\text{std}}$ across books). Cluster coverage estimates Overton pluralism, while JSD estimates distributional pluralism.}
    \resizebox{\columnwidth}{!}{%
    \begin{tabular}{llccccc}
        \toprule
        & Setup & GPT 4.1 & Llama 3.1 8B & Gemini 2.5 & OLMo2 1B & OLMo2 7B \\
        \midrule
        \multirow{3}{*}[0.1cm]{\rotatebox[origin=c]{90}{\textbf{Cover.}}}
          & $t=0.7$  & $41.4_{{\pm9.5}}$ & $63.8_{{\pm9.2}}$ & $71.8_{{\pm8.9}}$ & $72.7_{{\pm10.4}}$ & $59.3_{{\pm10.9}}$ \\
          & $t=1.5$  & $51.9_{{\pm11.4}}$ & $76.1_{{\pm7.7}}$ & $76.9_{{\pm8.6}}$ & $83.8_{{\pm8.0}}$ & $74.4_{{\pm9.0}}$ \\
          & Personas & $65.5_{{\pm11.0}}$ & $92.2_{{\pm5.2}}$ & $97.9_{{\pm2.6}}$ & $75.6_{{\pm9.8}}$ & $69.1_{{\pm10.4}}$ \\
        \midrule
        \multirow{3}{*}[0.1cm]{\rotatebox[origin=c]{90}{\textbf{JSD$^2$}}}
          & $t=0.7$  & $0.33_{{\pm.07}}$ & $0.25_{{\pm.05}}$ & $0.22_{{\pm.05}}$ & $0.23_{{\pm.06}}$ & $0.27_{{\pm.06}}$ \\
          & $t=1.5$  & $0.30_{{\pm.06}}$ & $0.23_{{\pm.04}}$ & $0.21_{{\pm.05}}$ & $0.21_{{\pm.05}}$ & $0.24_{{\pm.05}}$ \\
          & Personas & $0.26_{{\pm.06}}$ & $0.17_{{\pm.04}}$ & $0.12_{{\pm.03}}$ & $0.23_{{\pm.06}}$ & $0.24_{{\pm.05}}$ \\
        \bottomrule
    \end{tabular}%
    }
    \label{tab:coverage_jsd}
\end{table}

\begin{table}[ht]
    \small
    \centering
    \caption{Intra-cluster semantic similarity and entropy (mean$_{\pm\text{std}}$ across books).}
    \resizebox{\columnwidth}{!}{%
    \begin{tabular}{llccccc}
        \toprule
        & Setup & GPT 4.1 & Llama 3.1 8B & Gemini 2.5 & OLMo2 1B & OLMo2 7B \\
        \midrule
        \multirow{4}{*}{\rotatebox[origin=c]{90}{\textbf{Cl. Sim.}}}
          & $t=0.7$  & $0.68_{{\pm.05}}$ & $0.56_{{\pm.05}}$ & $0.51_{{\pm.05}}$ & $0.46_{{\pm.04}}$ & $0.56_{{\pm.04}}$ \\
          & $t=1.5$  & $0.61_{{\pm.04}}$ & $0.47_{{\pm.04}}$ & $0.49_{{\pm.04}}$ & $0.38_{{\pm.03}}$ & $0.47_{{\pm.03}}$ \\
          & Personas & $0.46_{{\pm.04}}$ & $0.40_{{\pm.04}}$ & $0.30_{{\pm.04}}$ & $0.44_{{\pm.03}}$ & $0.50_{{\pm.05}}$ \\
        \cmidrule(l){2-7}
          & \multicolumn{6}{c}{\cellcolor{gray!20}Original: $0.30_{{\pm.04}}$} \\
        \midrule
        \multirow{4}{*}{\rotatebox[origin=c]{90}{\textbf{$H$ (bits)}}}
          & $t=0.7$  & $3.14_{{\pm.54}}$ & $3.65_{{\pm.65}}$ & $3.85_{{\pm.71}}$ & $3.74_{{\pm.64}}$ & $3.53_{{\pm.61}}$ \\
          & $t=1.5$  & $3.37_{{\pm.56}}$ & $3.73_{{\pm.69}}$ & $3.91_{{\pm.71}}$ & $3.86_{{\pm.65}}$ & $3.69_{{\pm.64}}$ \\
          & Personas & $3.59_{{\pm.60}}$ & $4.13_{{\pm.69}}$ & $4.49_{{\pm.77}}$ & $3.77_{{\pm.63}}$ & $3.68_{{\pm.63}}$ \\
        \cmidrule(l){2-7}
          & \multicolumn{6}{c}{\cellcolor{gray!20}Original: $5.02_{{\pm.92}}$} \\
        \midrule
        \multirow{4}{*}{\rotatebox[origin=c]{90}{\textbf{$\hat{H}$}}}
          & $t=0.7$  & $0.77_{{\pm.10}}$ & $0.77_{{\pm.10}}$ & $0.78_{{\pm.09}}$ & $0.76_{{\pm.09}}$ & $0.76_{{\pm.09}}$ \\
          & $t=1.5$  & $0.76_{{\pm.09}}$ & $0.74_{{\pm.09}}$ & $0.78_{{\pm.09}}$ & $0.75_{{\pm.08}}$ & $0.74_{{\pm.08}}$ \\
          & Personas & $0.75_{{\pm.08}}$ & $0.78_{{\pm.08}}$ & $0.83_{{\pm.08}}$ & $0.75_{{\pm.09}}$ & $0.76_{{\pm.09}}$ \\
        \cmidrule(l){2-7}
          & \multicolumn{6}{c}{\cellcolor{gray!20}Original: $0.92_{{\pm.10}}$} \\
        \bottomrule
    \end{tabular}}
    \label{tab:similarity_entropy}
\end{table}

\paragraph{Perspective level.}
\label{sec:perspective_results}

To estimate perspective-level coverage, we use metrics similar to the aspect-level ones: JSD, cluster coverage, and normalized entropy. We add a novel perspective-level metric -- \emph{perspective diversity} -- which measures the average number of perspective clusters covered in a set of reviews relative to those covered in the evaluation set. We report results of all metrics in \Cref{tab:perspective_results} for using the k-means clustering algorithm with $k=5$.
We find that perspective-level coverage is considerably lower across all models than aspect-level coverage, indicating that while constituent aspects are adequately represented in generated texts, generated perspectives remain largely homogeneous.
This homogeneity is best seen through the perspective diversity metric, which is considerably lower than in the human reviews from $R_\mathit{eval}$ covering nearly all perspective clusters, with LLM-generated reviews frequently mapping to only one or two perspective clusters.
These results show that while LLMs are capable of generating texts corresponding to diverse aspects, this diversity is merely performative, with the overall perspectives present in the reviews still largely \textit{monocultural}.

\begin{table}[ht]
    \centering
    \caption{JSD$^2$, normalised entropy, perspective diversity, and perspective coverage (mean$_{\pm\text{std}}$ across books).}
    \resizebox{\columnwidth}{!}{%
    \begin{tabular}{llccccc}
        \toprule
        & Setup & GPT 4.1 & Llama 3.1 8B & Gemini 2.5 & OLMo2 1B & OLMo2 7B \\
        \midrule
        \multirow{4}{*}{\rotatebox[origin=c]{90}{\textbf{Persp. Cov.}}}
          & Baseline & $35.9_{{\pm16.7}}$ & $31.7_{{\pm14.5}}$ & $28.5_{{\pm13.1}}$ & $43.4_{{\pm19.6}}$ & $38.4_{{\pm18.0}}$ \\
          & $t=1.5$  & $40.8_{{\pm18.2}}$ & $36.4_{{\pm16.2}}$ & $29.1_{{\pm12.8}}$ & $51.6_{{\pm20.4}}$ & $44.7_{{\pm18.2}}$ \\
          & Personas & $50.4_{{\pm19.8}}$ & $38.4_{{\pm17.6}}$ & $33.8_{{\pm13.3}}$ & $39.5_{{\pm17.8}}$ & $44.4_{{\pm19.3}}$ \\
        \cmidrule(l){2-7}
          & \multicolumn{6}{c}{\cellcolor{gray!20}Original: $97.6_{{\pm7.1}}$} \\
        \midrule
        \multirow{4}{*}{\rotatebox[origin=c]{90}{\textbf{JSD$^2$}}}
          & $t=0.7$  & $0.085_{{\pm.061}}$ & $0.080_{{\pm.046}}$ & $0.087_{{\pm.050}}$ & $0.070_{{\pm.039}}$ & $0.075_{{\pm.041}}$ \\
          & $t=1.5$  & $0.075_{{\pm.046}}$ & $0.077_{{\pm.041}}$ & $0.087_{{\pm.055}}$ & $0.063_{{\pm.037}}$ & $0.070_{{\pm.038}}$ \\
          & Personas & $0.065_{{\pm.034}}$ & $0.071_{{\pm.036}}$ & $0.075_{{\pm.036}}$ & $0.073_{{\pm.038}}$ & $0.071_{{\pm.039}}$ \\
        \cmidrule(l){2-7}
          & \multicolumn{6}{c}{\cellcolor{gray!20}Original: ---} \\
        \midrule
        \multirow{4}{*}{\rotatebox[origin=c]{90}{\textbf{Norm. $\hat{H}$}}}
          & $t=0.7$  & $0.218_{{\pm.290}}$ & $0.204_{{\pm.308}}$ & $0.146_{{\pm.282}}$ & $0.247_{{\pm.244}}$ & $0.233_{{\pm.289}}$ \\
          & $t=1.5$  & $0.251_{{\pm.284}}$ & $0.244_{{\pm.316}}$ & $0.145_{{\pm.271}}$ & $0.299_{{\pm.241}}$ & $0.283_{{\pm.291}}$ \\
          & Personas & $0.305_{{\pm.257}}$ & $0.212_{{\pm.271}}$ & $0.175_{{\pm.253}}$ & $0.235_{{\pm.259}}$ & $0.252_{{\pm.268}}$ \\
        \cmidrule(l){2-7}
          & \multicolumn{6}{c}{\cellcolor{gray!20}Original: $0.496_{{\pm.134}}$} \\
        \midrule
        \multirow{4}{*}{\rotatebox[origin=c]{90}{\textbf{Persp. Div.}}}
          & Baseline & $1.74_{{\pm.79}}$ & $1.54_{{\pm.70}}$ & $1.38_{{\pm.62}}$ & $2.11_{{\pm.96}}$ & $1.87_{{\pm.88}}$ \\
          & $t=1.5$  & $1.99_{{\pm.88}}$ & $1.77_{{\pm.79}}$ & $1.41_{{\pm.60}}$ & $2.52_{{\pm.99}}$ & $2.19_{{\pm.90}}$ \\
          & Personas & $2.46_{{\pm.97}}$ & $1.86_{{\pm.82}}$ & $1.64_{{\pm.63}}$ & $1.92_{{\pm.84}}$ & $2.15_{{\pm.91}}$ \\
        \cmidrule(l){2-7}
          & \multicolumn{6}{c}{\cellcolor{gray!20}Original: $4.88_{{\pm.36}}$} \\
        \bottomrule
    \end{tabular}%
    }
    \label{tab:perspective_results}
\end{table}
\section{Analysis}
\label{sec:analysis}

We now analyse the validity of our framework by evaluating cluster contents and quality and approximating the proportion of reviews present in model pre-training sets. We first categorize the aspect clusters into meaningful topic categories and analyse their coverage (\sect{sec:topic_evaluation}), then evaluate the coherence and separability of the aspect clusters (\sect{sec:cluster_coherence}), and finally analyse the pre-training corpus of the \texttt{OLMo} model family (\sect{sec:memorization}).

\subsection{Topic evaluation}
\label{sec:topic_evaluation}

Results of our experiments (\sect{sec:results}) have identified the pluralistic gap, where the LLM-generated reviews lack diversity of perspectives. We now aim to identify which aspects and perspectives are consistently underrepresented. To do this, we label each inferred aspect cluster with a category and corresponding sentiment. For each of the aspect clusters across the analysis subset, we use five sentences closest to each aspect centroid to label clusters using \texttt{GPT-4.1-mini}. For labels, we use a predefined taxonomy of 24 categories proposed by \citet{yang2025matters} as a fixed set,
making labels consistent and directly comparable across books. The categories span both more objective (plot, characters, writing style) and subjective categories (emotional impact, enjoyment, and expectation fulfillment). We also assign a sentiment label (positive, negative, neutral, or mixed) expressed in the aspect toward that category. The resulting labels 
allow us to pinpoint the reasons LLMs diverge from human reviews in the topic space, and exactly along which aspects.

\begin{figure}[ht]
    \centering
    \begin{subfigure}[b]{\columnwidth}
        \includegraphics[width=\linewidth]{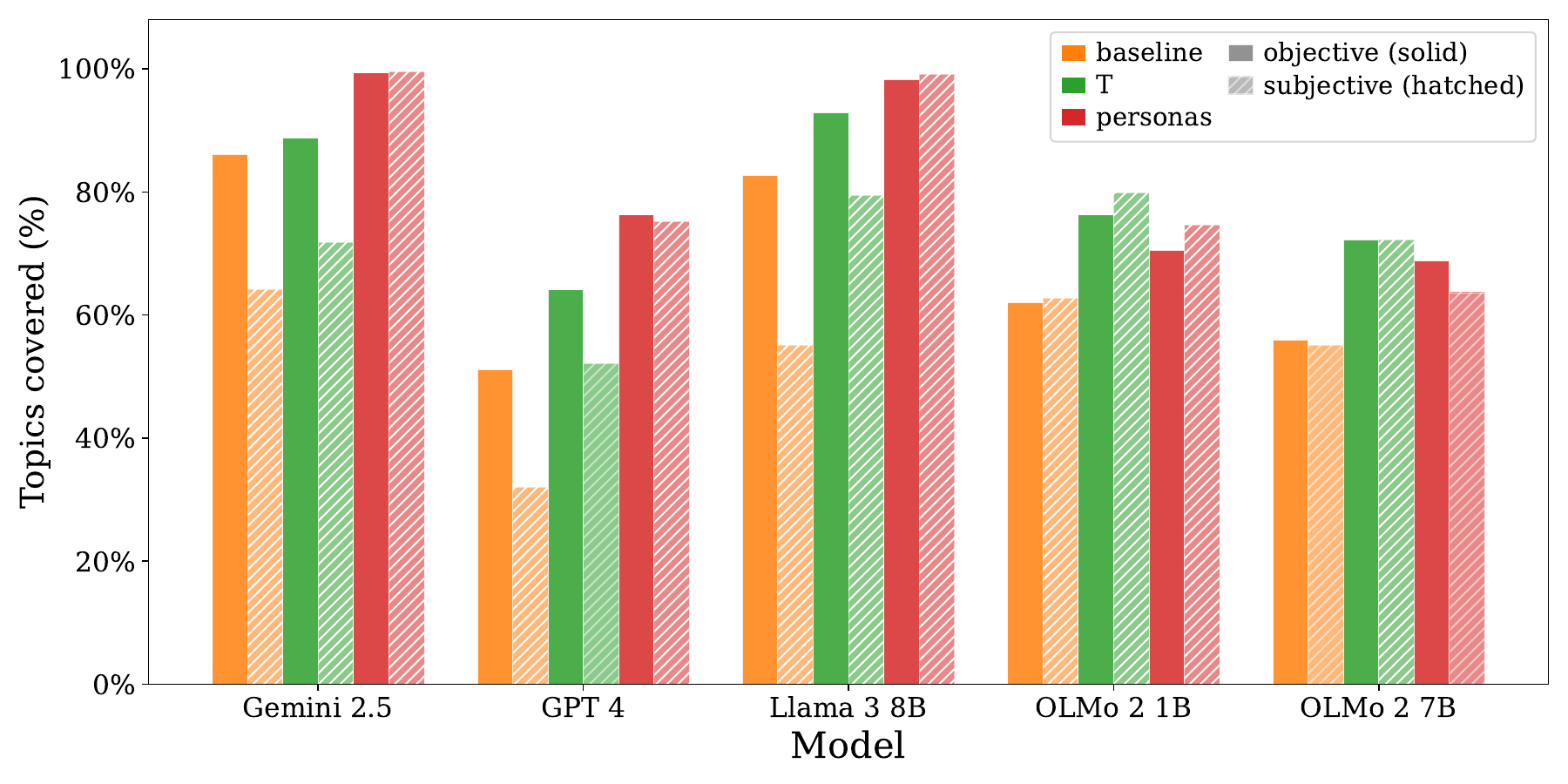}
        \caption{Objective vs. subjective coverage.}
        \label{fig:aspect_coverage}
    \end{subfigure}

    \vspace{0.5em}

    \begin{subfigure}[b]{\columnwidth}
        \includegraphics[width=\linewidth]{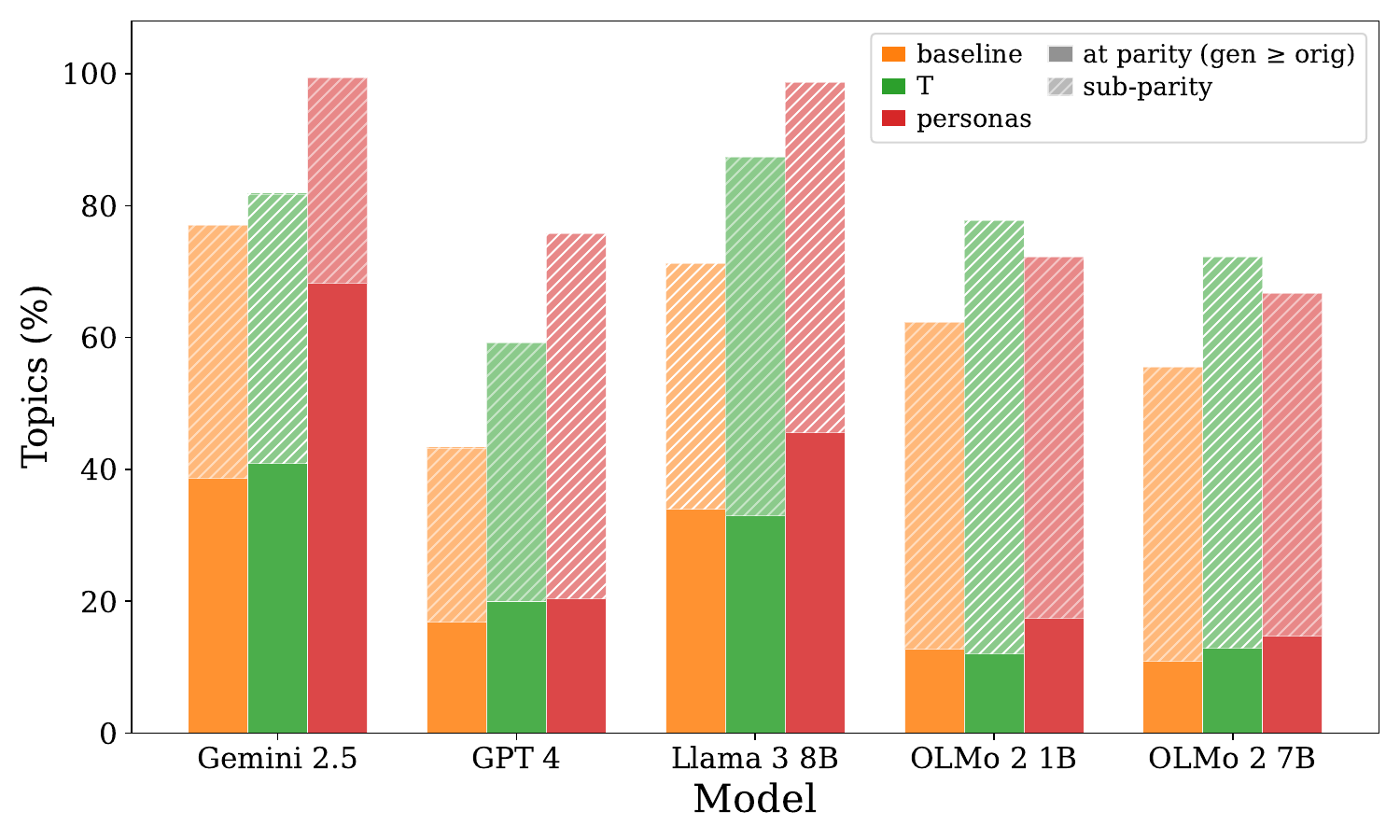}
        \caption{Parity of coverage across models.}
        \label{fig:coverage_parity}
    \end{subfigure}

    \caption{Topic coverage across generation modes for objective and subjective categories (\ref{fig:aspect_coverage}), and parity of aspect coverage compared to original distribution (\ref{fig:coverage_parity}).}
    \label{fig:coverage_combined}
\end{figure}

To jointly evaluate the coverage of aspects across books and study the influence of prompting configurations, we aggregate aspects into subjective and objective categories, then measure the percentage of those aspects covered across configurations and models. In \Cref{fig:aspect_coverage}, we report the difference in coverage between objective (full bar) and subjective (shaded bar) aspects. We observe that across models, persona prompting generally affects the coverage of subjective aspects, even when it does not achieve maximum coverage (as with \texttt{GPT 4.1}). In OLMo models, higher temperature has a stronger influence on topic coverage than persona prompting. Finally, model size does not necessarily imply lower coverage, as \texttt{OLMo2 1B} generally achieves higher coverage than \texttt{GPT 4.1}.

To complement Overton pluralism measured by topic coverage (1 review per topic at least), we evaluate on distributional pluralism, which is more informative. In \Cref{fig:coverage_parity}, we compare whether the distributions of the LLM-generated topics correspond to human ones. These results indicate that Overton pluralism does not guarantee distributional pluralism, showing models report more spurious rather than systematic coverage of diverse topics.

\subsection{Cluster coherence}
\label{sec:cluster_coherence}

We now aim to verify the validity of the produced cluster assignments. 
We evaluate aspect-cluster coherence using a leave-one-out intrinsic probe: for each cluster, we sample four representative sentences and one intruder sentence from a different cluster. Then, we use \texttt{GPT 4.1} for LLM-as-a-judge to determine the intruder. We vary this setup across two dimensions: (1) whether representative sentences are drawn from the centroid or sampled randomly, and (2) whether the outlier comes from a random or the closest neighbouring cluster. The LLM-as-a-judge prompt is provided in Appendix~\ref{sec:prompts}. 

We report the results in \Cref{tab:topic_coherence}. Intruder detection accuracy ranges from $72.4\%$ in the hardest configuration (random--closest) to $94.6\%$ in the easiest (centroid--random), against a $20\%$ random baseline, confirming that the inferred topic structure is coherent and separable.

\begin{table}[h]
\caption{Leave-one-out topic coherence accuracy across representative document and outlier selection strategies (baseline = 20.0\%).}
\centering
\begin{tabular}{lcc}
\toprule
 & $\text{outlier}_{\text{random}}$ & $\text{outlier}_{\text{closest}}$ \\
\midrule
$\text{in-cluster}_{\text{centroid}}$ & $94.6\%$ & $94.3\%$ \\
$\text{in-cluster}_{\text{random}}$   & $77.4\%$ & $72.4\%$ \\
\bottomrule
\end{tabular}
\label{tab:topic_coherence}
\end{table}

For the qualitative analysis, we manually reviewed the errors across all setups. In \Cref{tab:outlier-examples} we show one correct example alongside two error cases that illustrate different failure modes: one case where the mistake originated from the LLM's identification, and another case where an example was incorrectly flagged as an outlier due to a clustering mismatch, as evidenced by its sentiment being opposite to that of the other examples in the cluster, while focused around the same topic.

\begin{table*}[ht!]
\centering
\scriptsize
\renewcommand{\arraystretch}{1.1}
\begin{tabularx}{\textwidth}{@{} p{1.8cm} X p{0.7cm} p{0.7cm} @{}}
\toprule
\textbf{Category} & \textbf{Example} & \centering\textbf{Pick} & \centering\arraybackslash\textbf{Outlier} \\
\midrule
\multirow{5}{*}{\textbf{Positive}} & \ttfamily I read somewhere that she writes a book every 10 years. & & \\
 & \ttfamily The first three hundred or so pages were absolutely riveting. & \centering\checkmark & \centering\arraybackslash\checkmark \\
 & \ttfamily I only wish that she had more books out! & & \\
 & \ttfamily She almost fell apart from the overwhelming responses, which were both good and bad. & & \\
 & \ttfamily It takes her forever to write a book (this one took 8 years). & & \\
\midrule
\multirow{5}{*}{\textbf{LLM Error}} & \ttfamily It did not make my tiny brain inferior. & \centering\checkmark & \\
 & \ttfamily I will say my initial impression was that's not true to life. & & \\
 & \ttfamily This turned out to be a dud. & & \\
 & \ttfamily The first three hundred or so pages were absolutely riveting. & & \centering\arraybackslash\checkmark \\
 & \ttfamily What a disappointment. & & \\
\midrule
\multirow{5}{*}{\parbox{2.4cm}{\textbf{Clustering Error}}} & \ttfamily This is on a bunch of Top Read lists, and I'm not really sure why. & \centering\checkmark & \\
 & \ttfamily This book was on the list of '100 Books You Must Read in Your Lifetime -- NY Times' which I am working on. & & \\
 & \ttfamily Definitely worth reading, I will probably read it again sometime in print because I did enjoy the book. & & \\
 & \ttfamily Read this book before you die. & & \\
 & \ttfamily I thoroughly enjoyed it and count it among my favorites of 2012. & & \centering\arraybackslash\checkmark \\
\bottomrule
\end{tabularx}
\caption{Leave-one-out detection examples for identifying cluster coherence, ``Pick'' shows the assigned, and ``Outlier'' the true outlier. We present a positive example of outlier identification, as well as two failure modes due to classification LLM error and clustering error due to a mismatch in sentiment of cluster constituents.}
\label{tab:outlier-examples}
\end{table*}

\subsection{Memorization}
\label{sec:memorization}
In this section, we aim to estimate the proportion of review data present in model pre-training data, in order to confirm LLMs were exposed to a review pool large enough for diversity.
We conduct two complementary analyses of the \texttt{DCLM} corpus \citep{li2024datacomplm}, which is used in the pre-training pipeline of \texttt{OLMo} models \citep{olmo20252olmo2furious}, and likely contributes to the training mixtures of other LLMs as well.
In both analyses, we first retrieve all URLs corresponding to Goodreads from DCLM, as well as alternative book-review sources determined from manual search and LLM suggestions, indicating the presence of the sources in the training set. The sources contain user-generated literary reviews, book overviews, or related discussion content. The first analysis focuses specifically on Goodreads-derived content within DCLM, while the second examines alternative book-review and reading-community websites represented in the crawl.

\paragraph{Goodreads.} The Goodreads subset of the DCLM crawl contains $25,831$ pages from two page types: $19,052$ individual review pages (\texttt{goodreads.com/review/show/}) and $6,779$ book-summary pages (\texttt{goodreads.com/book/show/}). Because each \texttt{review/show} page contains one complete review, these pages contribute $19,052$ reviews directly. The \texttt{book/show} pages additionally embed up to $30$ community reviews each; using truncation markers as a conservative estimate yields approximately $131,441$ further review texts, for a total estimate of $150,493$ Goodreads reviews. Linking crawl pages to the \texttt{goodreads/books} metadata via Goodreads IDs or normalized titles matched $14,873$ distinct books containing $16.1$ million Goodreads reviews in total in the dataset. The crawl captured approximately $131,971$ of these reviews, corresponding to an overall coverage rate of $0.82\%$. The coverage is highly skewed: $32.6\%$ of matched books have more than $5\%$ of their Goodreads reviews present in the crawl, while $12.5\%$ have coverage below $0.1\%$. Among the $100$ selected books used for downstream analysis, $77$ appear in the crawl, yielding $612$ matched \texttt{review/show} pages. 
The high number of Goodreads reviews present in the training data confirms the LLMs were exposed to community reviews, validating they could feasibly reproduce the diverse perspectives present in those reviews, if properly aligned.

\paragraph{Other sources.} To complement the Goodreads-focused analysis, we additionally examine review and discussion content originating from alternative reading communities and book-review platforms present in the DCLM crawl.
We use LLM-as-a-judge on a random sample of content from $100$ URLs of chosen book-related domains, estimating whether and how many book reviews and book overviews were present in the page. We then multiply the average number of reviews and overviews by the total number of pages for that domain present in the crawl. We report complete results in Appendix~\ref{sec:pre-training}. In total, the estimate comes to over $300k$ reviews and over $340k$ overviews, confirming that LLMs were exposed to a wide amount of evaluative book content.

\section{Conclusion} 




In our work, we present a framework for evaluating pluralism in LLMs by first identifying constituent aspects of perspectives from free-form text and then grouping them into perspective representations. This two-level scheme extends beyond previously used aggregate statistics and allows for a fine-grained analysis of the \textit{pluralistic gap} between human and LLM generations. We propose one concrete instance of our framework and apply it to a dataset of book reviews, extracting aspect and perspective distributions and allowing for comparison between authentic human reviews and LLM-generated ones. We verify results from previous works, which indicate that LLMs are part of a \textit{generative monoculture}, but also show that the pluralistic gap can be bridged by utilizing prompting techniques such as persona prompting. Our subsequent analyses verify the validity of our method through cluster coherence and separability and identify the coverage of which topics was improved. Taken together, we offer a principled way of analysing fine-grained diversity of perspectives in free-form texts, which can be used to identify concrete targets for pluralistic alignment.

\section*{Impact Statement}

In this work, we study pluralism in LLMs through fine-grained analysis of aspects that are constituent of human perspectives. We identify the \textit{pluralistic gap} and evaluate whether it can be bridged using simple prompting techniques. As LLM-generated content becomes more prevalent and humans interact with LLMs to a greater degree, it is paramount to fairly represent individual voices, no matter how infrequent they may be. Concerningly, various works show that LLMs sharpen the distribution of perspectives in the training data, either not encoding or not generating less frequent ones. Our work provides a fine-grained, structured framework for analysing this phenomenon and paves the way towards effective mitigation strategies aimed at mitigating the pluralistic gap.

\bibliography{sources}

@inproceedings{roadmap2024sorensen,
author = {Sorensen, Taylor and Moore, Jared and Fisher, Jillian and Gordon, Mitchell and Mireshghallah, Niloofar and Rytting, Christopher Michael and Ye, Andre and Jiang, Liwei and Lu, Ximing and Dziri, Nouha and Althoff, Tim and Choi, Yejin},
title = {Position: a roadmap to pluralistic alignment},
year = {2024},
publisher = {JMLR.org},
booktitle = {Proceedings of the 41st International Conference on Machine Learning},
articleno = {1882},
numpages = {23},
location = {Vienna, Austria},
series = {ICML'24}
}

@inproceedings{
wu2025generative,
title={Generative Monoculture in Large Language Models},
author={Fan Wu and Emily Black and Varun Chandrasekaran},
booktitle={The Thirteenth International Conference on Learning Representations},
year={2025},
url={https://openreview.net/forum?id=yZ7sn9pyqb}
}

@inproceedings{goodreads2018recommender,
  author       = {Mengting Wan and
                  Julian J. McAuley},
  editor       = {Sole Pera and
                  Michael D. Ekstrand and
                  Xavier Amatriain and
                  John O'Donovan},
  title        = {Item recommendation on monotonic behavior chains},
  booktitle    = {Proceedings of the 12th {ACM} Conference on Recommender Systems, RecSys
                  2018, Vancouver, BC, Canada, October 2-7, 2018},
  pages        = {86--94},
  publisher    = {{ACM}},
  year         = {2018},
  url          = {https://doi.org/10.1145/3240323.3240369},
  doi          = {10.1145/3240323.3240369},
  bibsource    = {dblp computer science bibliography, https://dblp.org}
}

@misc{russo2025pluralisticmoralgapunderstanding,
      title={The Pluralistic Moral Gap: Understanding Judgment and Value Differences between Humans and Large Language Models}, 
      author={Giuseppe Russo and Debora Nozza and Paul Röttger and Dirk Hovy},
      year={2025},
      eprint={2507.17216},
      archivePrefix={arXiv},
      primaryClass={cs.CL},
      url={https://arxiv.org/abs/2507.17216}, 
}

@misc{peeperkorn2025gap,
      title={Mind the Gap: Conformative Decoding to Improve Output Diversity of Instruction-Tuned Large Language Models}, 
      author={Max Peeperkorn and Tom Kouwenhoven and Dan Brown and Anna Jordanous},
      year={2025},
      eprint={2507.20956},
      archivePrefix={arXiv},
      primaryClass={cs.CL},
      url={https://arxiv.org/abs/2507.20956}, 
}

@inproceedings{jimenez2024swebench,
  author       = {Carlos E. Jimenez and
                  John Yang and
                  Alexander Wettig and
                  Shunyu Yao and
                  Kexin Pei and
                  Ofir Press and
                  Karthik R. Narasimhan},
  title        = {SWE-bench: Can Language Models Resolve Real-world {Github} Issues?},
  booktitle    = {The Twelfth International Conference on Learning Representations,
                  {ICLR} 2024, Vienna, Austria, May 7-11, 2024},
  publisher    = {OpenReview.net},
  year         = {2024},
  url          = {https://openreview.net/forum?id=VTF8yNQM66},
  bibsource    = {dblp computer science bibliography, https://dblp.org}
}

@article{phan2025humanity,
  author       = {Long Phan and
                  Alice Gatti and
                  Ziwen Han and
                  Nathaniel Li and
                  Josephina Hu and
                  Hugh Zhang and
                  Chen Bo Calvin Zhang and
                  Mohamed Shaaban and
                  John Ling and
                  Sean Shi and
                  Michael Choi and
                  Anish Agrawal and
                  Arnav Chopra and
                  Adam Khoja and
                  Ryan Kim and
                  Richard Ren and
                  Jason Hausenloy and
                  Oliver Zhang and
                  Mantas Mazeika and
                  Summer Yue and
                  Alexandr Wang and
                  Dan Hendrycks},
  title        = {Humanity's Last Exam},
  journal      = {CoRR},
  volume       = {abs/2501.14249},
  year         = {2025},
  url          = {https://doi.org/10.48550/arXiv.2501.14249},
  doi          = {10.48550/ARXIV.2501.14249},
  eprinttype   = {arXiv},
  eprint       = {2501.14249},
  bibsource    = {dblp computer science bibliography, https://dblp.org}
}

@inproceedings{
jiang2025artificial,
title={Artificial Hivemind: The Open-Ended Homogeneity of Language Models (and Beyond)},
author={Liwei Jiang and Yuanjun Chai and Margaret Li and Mickel Liu and Raymond Fok and Nouha Dziri and Yulia Tsvetkov and Maarten Sap and Yejin Choi},
booktitle={The Thirty-ninth Annual Conference on Neural Information Processing Systems Datasets and Benchmarks Track},
year={2025},
url={https://openreview.net/forum?id=saDOrrnNTz}
}

@misc{olmo20252olmo2furious,
      title={2 OLMo 2 Furious}, 
      author={Team OLMo and Pete Walsh and Luca Soldaini and Dirk Groeneveld and Kyle Lo and Shane Arora and Akshita Bhagia and Yuling Gu and Shengyi Huang and Matt Jordan and Nathan Lambert and Dustin Schwenk and Oyvind Tafjord and Taira Anderson and David Atkinson and Faeze Brahman and Christopher Clark and Pradeep Dasigi and Nouha Dziri and ... and Hannaneh Hajishirzi},
      year={2025},
      eprint={2501.00656},
      archivePrefix={arXiv},
      primaryClass={cs.CL},
      url={https://arxiv.org/abs/2501.00656}, 
}

@misc{zhang2025f2llmproc,
      title={F2LLM Technical Report: Matching SOTA Embedding Performance with 6 Million Open-Source Data}, 
      author={Ziyin Zhang and Zihan Liao and Hang Yu and Peng Di and Rui Wang},
      year={2025},
      eprint={2510.02294},
      archivePrefix={arXiv},
      primaryClass={cs.CL},
      url={https://arxiv.org/abs/2510.02294}, 
}

@article{grootendorst2022bertopic,
  title={BERTopic: Neural topic modeling with a class-based TF-IDF procedure},
  author={Grootendorst, Maarten},
  journal={arXiv preprint arXiv:2203.05794},
  year={2022}
}

@misc{ge2025personas,
      title={Scaling Synthetic Data Creation with 1,000,000,000 Personas}, 
      author={Tao Ge and Xin Chan and Xiaoyang Wang and Dian Yu and Haitao Mi and Dong Yu},
      year={2025},
      eprint={2406.20094},
      archivePrefix={arXiv},
      primaryClass={cs.CL},
      url={https://arxiv.org/abs/2406.20094}, 
}

@InProceedings{HDBSCAN,
author="Campello, Ricardo J. G. B.
and Moulavi, Davoud
and Sander, Joerg",
editor="Pei, Jian
and Tseng, Vincent S.
and Cao, Longbing
and Motoda, Hiroshi
and Xu, Guandong",
title="Density-Based Clustering Based on Hierarchical Density Estimates",
booktitle="Advances in Knowledge Discovery and Data Mining",
year="2013",
publisher="Springer Berlin Heidelberg",
address="Berlin, Heidelberg",
pages="160--172",
isbn="978-3-642-37456-2"
}

@inproceedings{
durmus2024towards,
title={Towards Measuring the Representation of Subjective Global Opinions in Language Models},
author={Esin Durmus and Karina Nguyen and Thomas Liao and Nicholas Schiefer and Amanda Askell and Anton Bakhtin and Carol Chen and Zac Hatfield-Dodds and Danny Hernandez and Nicholas Joseph and Liane Lovitt and Sam McCandlish and Orowa Sikder and Alex Tamkin and Janel Thamkul and Jared Kaplan and Jack Clark and Deep Ganguli},
booktitle={First Conference on Language Modeling},
year={2024},
url={https://openreview.net/forum?id=zl16jLb91v}
}

@misc{adams2025steerable,
      title={Steerable Pluralism: Pluralistic Alignment via Few-Shot Comparative Regression}, 
      author={Jadie Adams and Brian Hu and Emily Veenhuis and David Joy and Bharadwaj Ravichandran and Aaron Bray and Anthony Hoogs and Arslan Basharat},
      year={2025},
      eprint={2508.08509},
      archivePrefix={arXiv},
      primaryClass={cs.CL},
      url={https://arxiv.org/abs/2508.08509}, 
}

@inproceedings{chen2025spica,
    title = "{SPICA}: Retrieving Scenarios for Pluralistic In-Context Alignment",
    author = "Chen, Quan Ze  and
      Feng, Kevin  and
      Park, Chan Young  and
      Zhang, Amy X",
    editor = "Che, Wanxiang  and
      Nabende, Joyce  and
      Shutova, Ekaterina  and
      Pilehvar, Mohammad Taher",
    booktitle = "Findings of the Association for Computational Linguistics: ACL 2025",
    month = jul,
    year = "2025",
    address = "Vienna, Austria",
    publisher = "Association for Computational Linguistics",
    url = "https://aclanthology.org/2025.findings-acl.41/",
    doi = "10.18653/v1/2025.findings-acl.41",
    pages = "748--765",
    ISBN = "979-8-89176-256-5"
}

@article{grattafiori2024llama,
  title={The {Llama} 3 herd of models},
  author={Grattafiori, Aaron and Dubey, Abhimanyu and Jauhri, Abhinav and Pandey, Abhinav and Kadian, Abhishek and Al-Dahle, Ahmad and Letman, Aiesha and Mathur, Akhil and Schelten, Alan and Vaughan, Alex and others},
  journal={arXiv preprint arXiv:2407.21783},
  year={2024}
}

@inproceedings{vitsakis-etal-2024-voices,
    title = "Voices in a Crowd: Searching for clusters of unique perspectives",
    author = "Vitsakis, Nikolas  and
      Parekh, Amit  and
      Konstas, Ioannis",
    editor = "Al-Onaizan, Yaser  and
      Bansal, Mohit  and
      Chen, Yun-Nung",
    booktitle = "Proceedings of the 2024 Conference on Empirical Methods in Natural Language Processing",
    month = nov,
    year = "2024",
    address = "Miami, Florida, USA",
    publisher = "Association for Computational Linguistics",
    url = "https://aclanthology.org/2024.emnlp-main.696/",
    doi = "10.18653/v1/2024.emnlp-main.696",
    pages = "12517--12539"
}

@inproceedings{ajjour-etal-2019-modeling,
    title = "Modeling Frames in Argumentation",
    author = "Ajjour, Yamen  and
      Alshomary, Milad  and
      Wachsmuth, Henning  and
      Stein, Benno",
    editor = "Inui, Kentaro  and
      Jiang, Jing  and
      Ng, Vincent  and
      Wan, Xiaojun",
    booktitle = "Proceedings of the 2019 Conference on Empirical Methods in Natural Language Processing and the 9th International Joint Conference on Natural Language Processing (EMNLP-IJCNLP)",
    month = nov,
    year = "2019",
    address = "Hong Kong, China",
    publisher = "Association for Computational Linguistics",
    url = "https://aclanthology.org/D19-1290/",
    doi = "10.18653/v1/D19-1290",
    pages = "2922--2932",
}

@misc{li2024datacomplm,
      title={DataComp-LM: In search of the next generation of training sets for language models}, 
      author={Jeffrey Li and Alex Fang and Georgios Smyrnis and Maor Ivgi and
        Matt Jordan and Samir Gadre and Hritik Bansal and Etash Guha and
        Sedrick Keh and Kushal Arora and Saurabh Garg and Rui Xin and
        Niklas Muennighoff and Reinhard Heckel and Jean Mercat and
        Mayee Chen and Suchin Gururangan and Mitchell Wortsman and
        Alon Albalak and others},
      year={2024},
      eprint={2406.11794},
      archivePrefix={arXiv},
      primaryClass={id='cs.LG' full_name='Machine Learning' is_active=True alt_name=None in_archive='cs' is_general=False description='Papers on all aspects of machine learning research (supervised, unsupervised, reinforcement learning, bandit problems, and so on) including also robustness, explanation, fairness, and methodology. cs.LG is also an appropriate primary category for applications of machine learning methods.'}
}

@inproceedings{reimers-etal-2019-classification,
    title = "Classification and Clustering of Arguments with Contextualized Word Embeddings",
    author = "Reimers, Nils  and
      Schiller, Benjamin  and
      Beck, Tilman  and
      Daxenberger, Johannes  and
      Stab, Christian  and
      Gurevych, Iryna",
    editor = "Korhonen, Anna  and
      Traum, David  and
      M{\`a}rquez, Llu{\'i}s",
    booktitle = "Proceedings of the 57th Annual Meeting of the Association for Computational Linguistics",
    month = jul,
    year = "2019",
    address = "Florence, Italy",
    publisher = "Association for Computational Linguistics",
    url = "https://aclanthology.org/P19-1054/",
    doi = "10.18653/v1/P19-1054",
    pages = "567--578",
}

@ARTICLE{basile2022perspective,   
AUTHOR={Basile, Valerio  and Caselli, Tommaso  and Balahur, Alexandra  and Ku, Lun-Wei },           
TITLE={Editorial: Bias, Subjectivity and Perspectives in Natural Language Processing},          
JOURNAL={Frontiers in Artificial Intelligence},         
VOLUME={Volume 5 - 2022}, 
YEAR={2022},  
URL={https://www.frontiersin.org/journals/artificial-intelligence/articles/10.3389/frai.2022.926435}, 
DOI={10.3389/frai.2022.926435}, 
ISSN={2624-8212}, 
}

@inproceedings{fleisig-etal-2024-perspectivist,
    title = "The Perspectivist Paradigm Shift: Assumptions and Challenges of Capturing Human Labels",
    author = "Fleisig, Eve  and
      Blodgett, Su Lin  and
      Klein, Dan  and
      Talat, Zeerak",
    editor = "Duh, Kevin  and
      Gomez, Helena  and
      Bethard, Steven",
    booktitle = "Proceedings of the 2024 Conference of the North American Chapter of the Association for Computational Linguistics: Human Language Technologies (Volume 1: Long Papers)",
    month = jun,
    year = "2024",
    address = "Mexico City, Mexico",
    publisher = "Association for Computational Linguistics",
    url = "https://aclanthology.org/2024.naacl-long.126/",
    doi = "10.18653/v1/2024.naacl-long.126",
    pages = "2279--2292"
}

@article{frenda2025perspectivist,
  title   = {Perspectivist approaches to natural language processing: a survey},
  author  = {Frenda, Simona and Abercrombie, Gavin and Basile, Valerio and others},
  journal = {Language Resources and Evaluation},
  volume  = {59},
  pages   = {1719--1746},
  year    = {2025},
  doi     = {10.1007/s10579-024-09766-4}
}

@misc{cambridge_perspective,
  title        = {Perspective},
  author       = {{Cambridge University Press}},
  year         = {n.d.},
  howpublished = {\url{https://dictionary.cambridge.org/dictionary/english/perspective}},
  note         = {Accessed: 2026-05-05}
}

@article{achiam2023gpt,
  title={Gpt-4 technical report},
  author={Achiam, Josh and Adler, Steven and Agarwal, Sandhini and Ahmad, Lama and Akkaya, Ilge and Aleman, Florencia Leoni and Almeida, Diogo and Altenschmidt, Janko and Altman, Sam and Anadkat, Shyamal and others},
  journal={arXiv preprint arXiv:2303.08774},
  year={2023}
}

@article{sanh2019distilbert,
  title={DistilBERT, a distilled version of BERT: smaller, faster, cheaper and lighter},
  author={Sanh, Victor and Debut, Lysandre and Chaumond, Julien and Wolf, Thomas},
  journal={arXiv preprint arXiv:1910.01108},
  year={2019}
}

@article{comanici2025gemini,
  title={Gemini 2.5: Pushing the frontier with advanced reasoning, multimodality, long context, and next generation agentic capabilities},
  author={Comanici, Gheorghe and Bieber, Eric and Schaekermann, Mike and Pasupat, Ice and Sachdeva, Noveen and Dhillon, Inderjit and Blistein, Marcel and Ram, Ori and Zhang, Dan and Rosen, Evan and others},
  journal={arXiv preprint arXiv:2507.06261},
  year={2025}
}

@inproceedings{yang2025matters,
    title = "What Matters in Evaluating Book-Length Stories? A Systematic Study of Long Story Evaluation",
    author = "Yang, Dingyi  and
      Jin, Qin",
    editor = "Che, Wanxiang  and
      Nabende, Joyce  and
      Shutova, Ekaterina  and
      Pilehvar, Mohammad Taher",
    booktitle = "Proceedings of the 63rd Annual Meeting of the Association for Computational Linguistics (Volume 1: Long Papers)",
    month = jul,
    year = "2025",
    address = "Vienna, Austria",
    publisher = "Association for Computational Linguistics",
    url = "https://aclanthology.org/2025.acl-long.799/",
    doi = "10.18653/v1/2025.acl-long.799",
    pages = "16375--16398",
    ISBN = "979-8-89176-251-0"
}

@article{McInnes2018, doi = {10.21105/joss.00861}, url = {https://doi.org/10.21105/joss.00861}, year = {2018}, publisher = {The Open Journal}, volume = {3}, number = {29}, pages = {861}, author = {McInnes, Leland and Healy, John and Saul, Nathaniel and Großberger, Lukas}, title = {UMAP: Uniform Manifold Approximation and Projection}, journal = {Journal of Open Source Software} }

@inproceedings{meister2025benchmarking,
    title = "Benchmarking Distributional Alignment of Large Language Models",
    author = "Meister, Nicole  and
      Guestrin, Carlos  and
      Hashimoto, Tatsunori",
    editor = "Chiruzzo, Luis  and
      Ritter, Alan  and
      Wang, Lu",
    booktitle = "Proceedings of the 2025 Conference of the Nations of the Americas Chapter of the Association for Computational Linguistics: Human Language Technologies (Volume 1: Long Papers)",
    month = apr,
    year = "2025",
    address = "Albuquerque, New Mexico",
    publisher = "Association for Computational Linguistics",
    url = "https://aclanthology.org/2025.naacl-long.2/",
    doi = "10.18653/v1/2025.naacl-long.2",
    pages = "24--49",
    ISBN = "979-8-89176-189-6"
}

@inproceedings{santurkar2023opinions,
  title={Whose opinions do language models reflect?},
  author={Santurkar, Shibani and Durmus, Esin and Ladhak, Faisal and Lee, Cinoo and Liang, Percy and Hashimoto, Tatsunori},
  booktitle={Proceedings of the 40th International Conference on Machine Learning},
  pages={29971--30004},
  year={2023}
}
\bibliographystyle{icml2026}

\newpage
\appendix
\onecolumn

\section{Prompts}
\label{sec:prompts}

We report all prompts used to generate reviews across various setups and LLM-as-a-judge for cluster coherence, labelling and pretraining data analysis in \Cref{tab:prompts_combined}.

\begin{table}[h!]
\centering
\caption{Prompts used for generating reviews (top), leave-one-out cluster coherence evaluation (second), cluster labelling (third), and web page classification (bottom).}
\resizebox{\textwidth}{!}{%
\begin{tabular}{@{}lll@{}}
\toprule
\textbf{Usage} & \textbf{Prompt Type} & \textbf{Content} \\
\midrule
\multirow{3}{*}{\parbox[c]{2cm}{\centering\textbf{Review Generation}}}
& system & \parbox[c]{0.7\textwidth}{\texttt{"You just finished reading a book."}} \\[2mm]
\cmidrule(l){2-3}
& baseline user & \parbox[c]{0.7\textwidth}{\texttt{"Write a book review for [TITLE] by [AUTHOR]. Provide an integer rating from 1 to 5, and a written review. Format it like 'Rating: X/5, Review:...'"}} \\[2mm]
\cmidrule(l){2-3}
& persona user & \parbox[c]{0.7\textwidth}{\texttt{"You are [PERSONA]. Write a book review for [TITLE] by [AUTHOR]. Provide an integer rating from 1 to 5, and a written review. Format it like 'Rating: X/5, Review:...'"}} \\[2mm]
\midrule
\multirow{3}{*}[1cm]{\parbox[c]{2cm}{\centering\textbf{Topic Coherence Evaluation}}}
& system & \parbox[c]{0.7\textwidth}{\texttt{"You are an expert at analyzing book reviews. You will receive 5 book review sentences. Four of them come from the same topic cluster and share a common theme or aspect. One is an outlier from a different cluster. Your task: identify which sentence (by number, 1--5) is the odd one out --- the one that does NOT belong with the other four. Respond with a JSON object only."}} \\[2mm]
\cmidrule(l){2-3}
& user & \parbox[c]{0.7\textwidth}{\texttt{"Five book review sentences: 1. [s1] ... 5. [s5]. Which sentence is the odd one out?"}} \\[2mm]
\cmidrule(l){2-3}
& response format & \parbox[c]{0.7\textwidth}{\texttt{\{"outlier": <int 1--5>, "reason": <string>\}}} \\[2mm]
\midrule
\multirow{3}{*}[1cm]{\parbox[c]{2cm}{\centering\textbf{Cluster Labelling}}}
& system & \parbox[c]{0.7\textwidth}{\texttt{"You are an expert at analyzing book reviews. You will receive 5 representative sentences from a single review topic cluster. Your task is to identify: 1. The aspect of the book being discussed --- pick the single best match from the provided enum. 2. The sentiment towards that aspect (positive, negative, or mixed). Base your judgment on what all 5 sentences have in common. Respond with a JSON object only."}} \\[2mm]
\cmidrule(l){2-3}
& aspects enum & \parbox[c]{0.7\textwidth}{\texttt{plot\_development, structure, ending, character\_development, characterization, character\_relationships, character\_diversity, writing\_style, language, readability, theme\_exploration, theme\_depth, world\_building, setting, empathy, emotional\_depth, enjoyment, engagement, genre\_fulfillment, premise\_fulfillment, originality, content\_warnings, cover\_design, personal\_bias}} \\[2mm]
\cmidrule(l){2-3}
& response format & \parbox[c]{0.7\textwidth}{\texttt{\{"aspect": <enum>, "sentiment": "positive" | "negative" | "mixed"\}}} \\[2mm]
\midrule
\multirow{2}{*}[1.2cm]{\parbox[c]{2cm}{\centering\textbf{Page Classification}}}
& user & \parbox[c]{0.7\textwidth}{\texttt{"You are a data quality analyst. I will give you the text content scraped from a web page. Your task: determine how many individual book reviews and book overviews are present. A `book review' evaluates or discusses a specific book (professional critic, reader review, or blog post). A `book overview OR excerpt' is a summary or recap of a book's plot/content without being evaluative --- e.g. publisher blurbs, plot summaries, `about the book' sections, editorial synopses, or excerpts. Do not flag movie or series reviews, or overviews on non-book topics. Page URL: [URL]. Page text (truncated): [TEXT]."}} \\[2mm]
\cmidrule(l){2-3}
& response format & \parbox[c]{0.7\textwidth}{\texttt{\{"review\_count": <int>, "has\_reviews": <bool>, "overview\_count": <int>, "has\_overviews": <bool>, "confidence": "high"|"medium"|"low", "notes": <one short sentence or empty string>\}}} \\
\bottomrule
\end{tabular}%
}
\label{tab:prompts_combined}
\end{table}

\section{Semantic Similarity}
\label{sec:semantic_similarity}

\begin{figure}[ht]
    \centering
    \begin{subfigure}[b]{0.48\textwidth}
        \includegraphics[width=\linewidth]{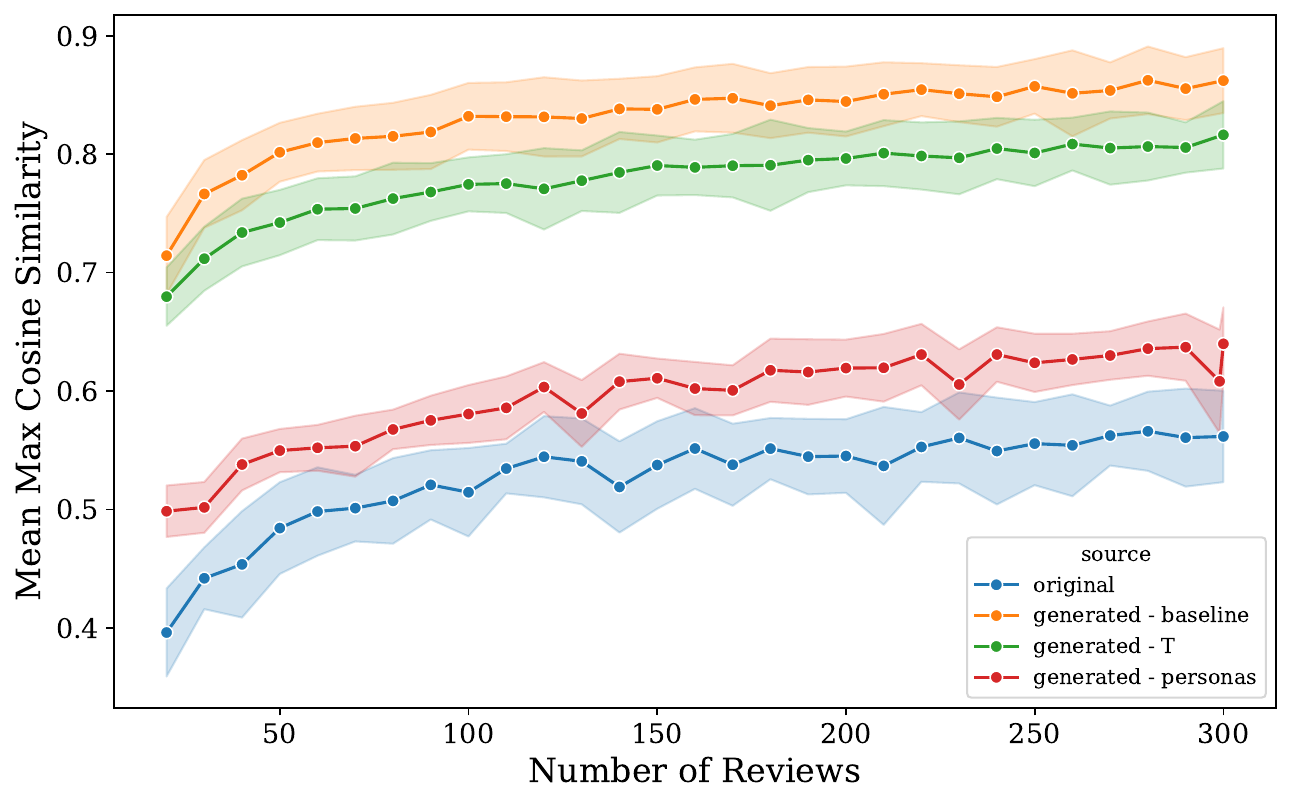}
        \caption{Gemini 2.5}
        \label{fig:app-sim-gem25}
    \end{subfigure}
    \hfill
    \begin{subfigure}[b]{0.48\textwidth}
        \includegraphics[width=\linewidth]{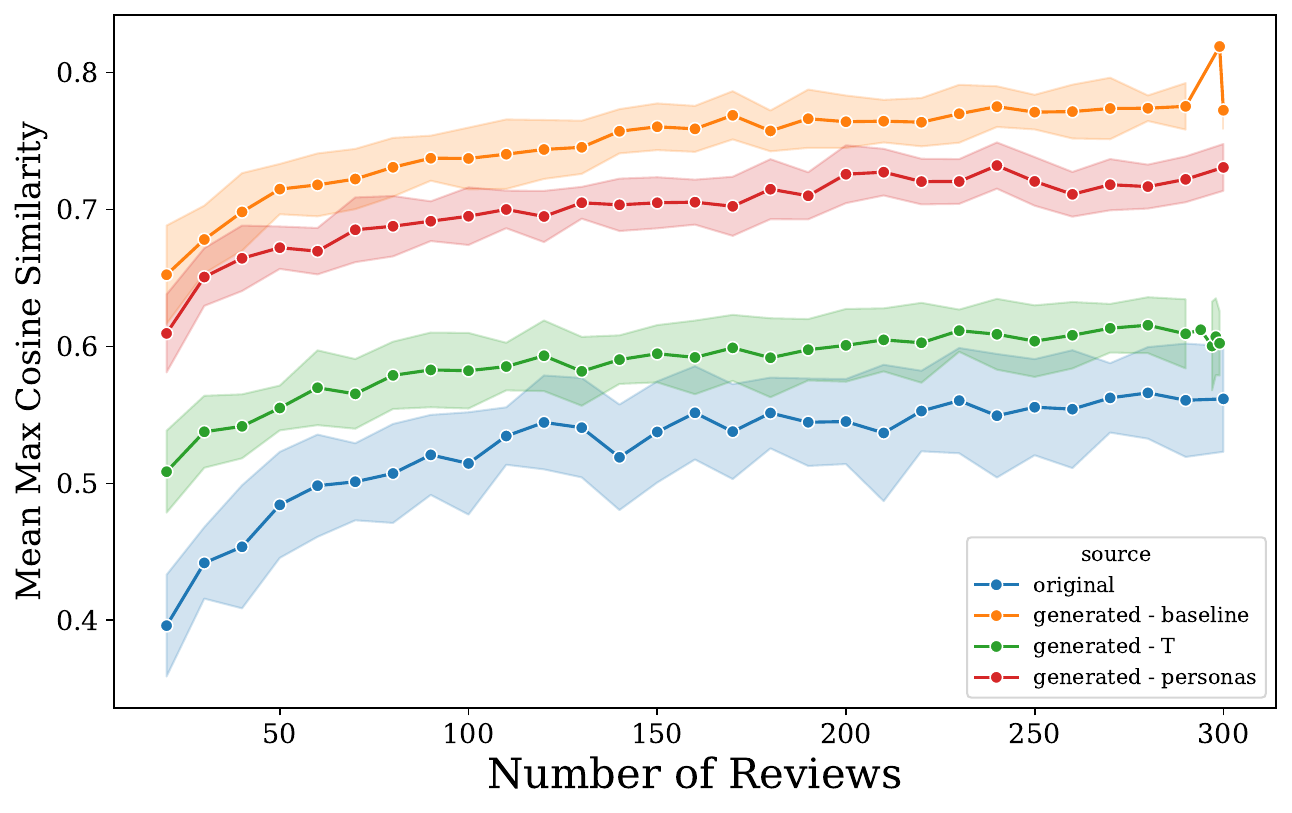}
        \caption{OLMo2 1b}
        \label{fig:app-sim-olmo1b}
    \end{subfigure}

    \vspace{0.5em}

    \begin{subfigure}[b]{0.48\textwidth}
        \includegraphics[width=\linewidth]{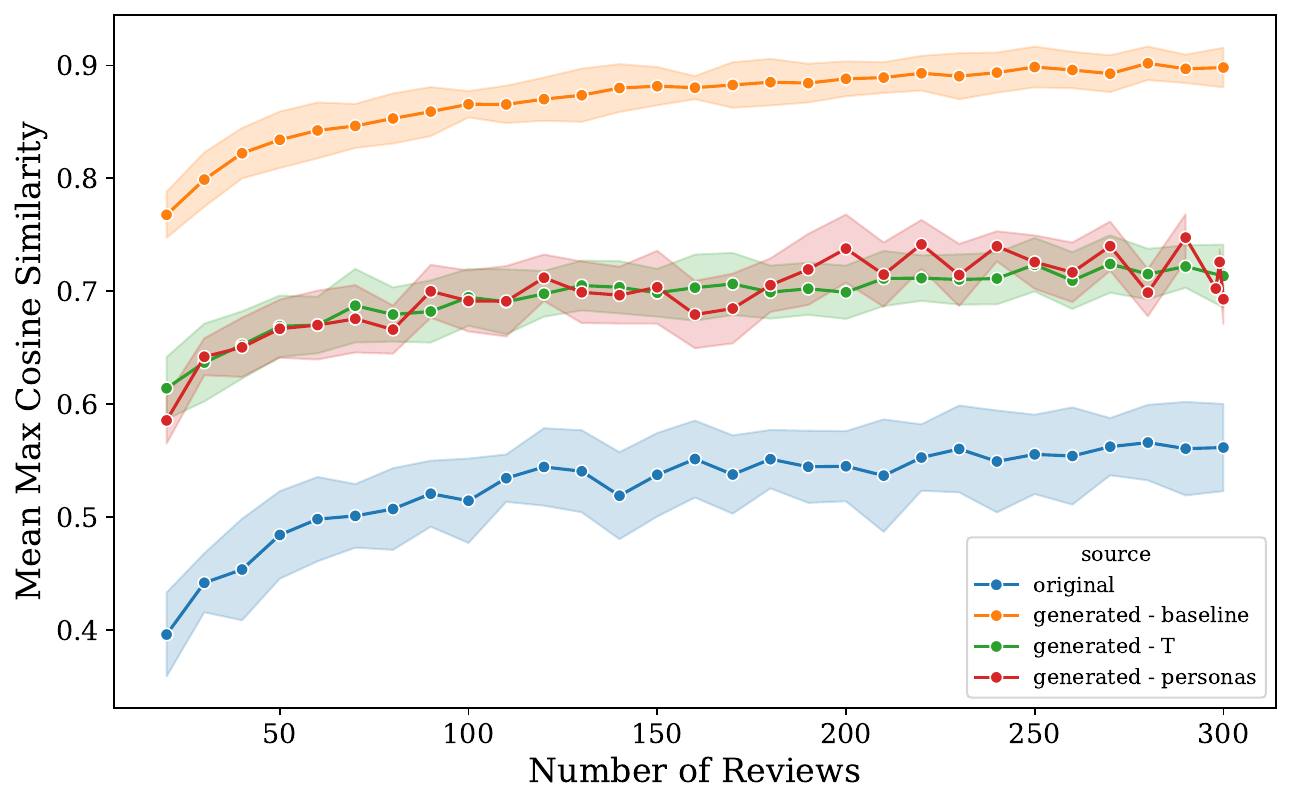}
        \caption{Llama 3.1 8B}
        \label{fig:app-sim-llama31}
    \end{subfigure}
    \hfill
    \begin{subfigure}[b]{0.48\textwidth}
        \includegraphics[width=\linewidth]{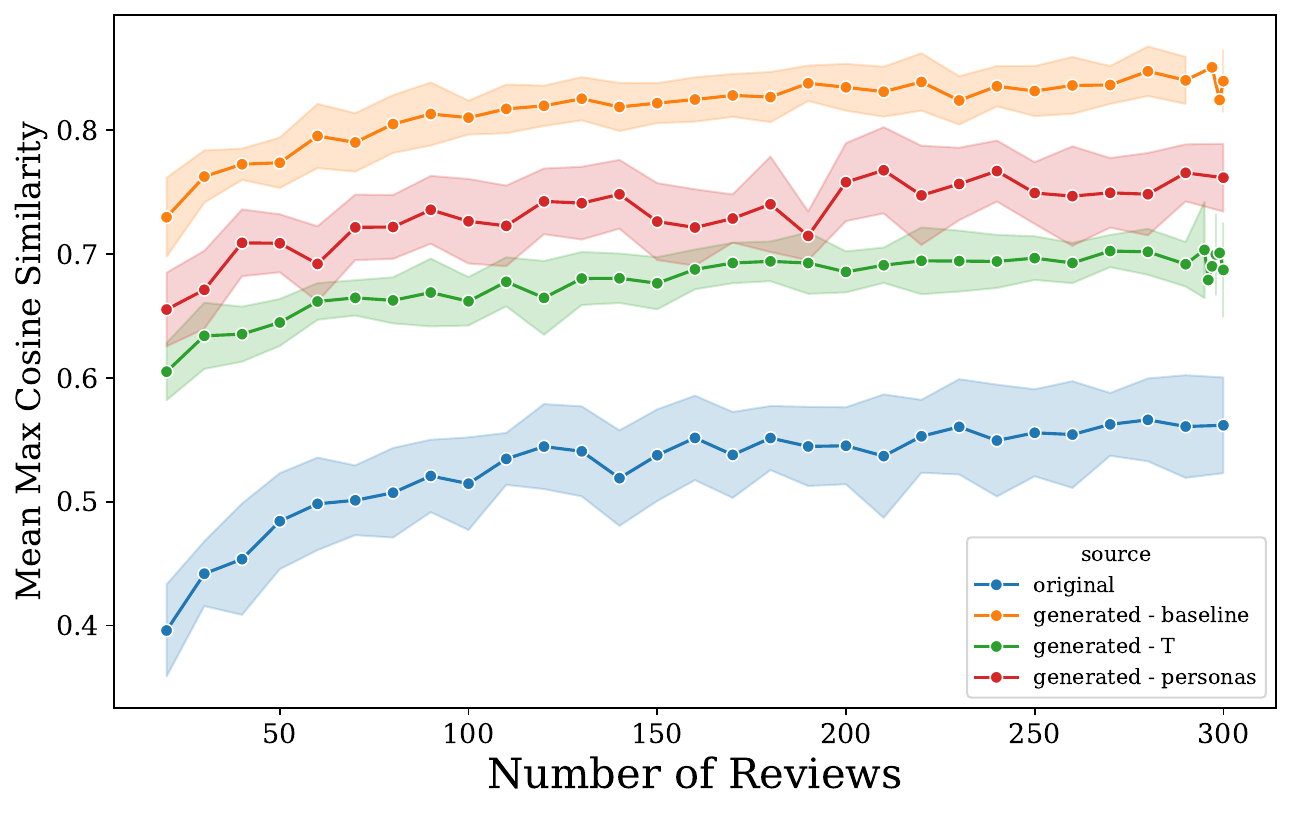}
        \caption{OLMo2 7b}
        \label{fig:app-sim-olmo7b}
    \end{subfigure}

    \caption{Similarity results across models.}
    \label{fig:similarity_results}
\end{figure}

We report the mean max semantic similarity of novel batches of reviews for other models used throughout our experiments (\Cref{fig:similarity}) in \Cref{fig:similarity_results}. We find that similar findings to ones for \texttt{GPT-4.1} hold for other models. The most notable cases are Gemini 2.5 (\Cref{fig:app-sim-gem25}), where persona prompting exhibits the strongest effect and LLaMA 3.1 (\Cref{fig:app-sim-llama31}), where persona prompting introduces little novel content compared to sampling with higher temperature. Overall, we believe (and the results support) the fact that properly utilizing personas in generation is an emergent capability in LLMs, as evidenced by semantic similarity of novel content being much higher for Gemini and GPT, the two largest models we studied.   

\section{Pre-training Dataset Analysis.}
\label{sec:pre-training}

We report the full results of reviews and Goodreads overview pages identified in the DCLM component of the OLMo pretraining data mix in \Cref{tab:book_sources}.

\begin{table}[ht]
    \small
    \centering
    \caption{Estimated review and overview pages per source (sources with $<$10\% in both categories omitted).}
    \resizebox{\textwidth}{!}{%
    \begin{tabular}{lrrrrrrr}
        \toprule
        & & \multicolumn{3}{c}{\textbf{Reviews}} & \multicolumn{3}{c}{\textbf{Overviews}} \\
        \cmidrule(lr){3-5} \cmidrule(lr){6-8}
        Source & Total Rows & \% Pages & Avg Count & Est. Pages & \% Pages & Avg Count & Est. Pages \\
        \midrule
        \texttt{amazon.com}                              & 155,879  & 26 & 1.03 & 40,529  & 13 & 0.13 & 20,264  \\
        \texttt{barnesandnoble.com}                      & 293,727  & 26 & 0.68 & 76,369  & 62 & 0.66 & 182,111 \\
        \texttt{bookpage.com}                            & 8,762    & 60 & 0.62 & 5,257   & 80 & 0.93 & 7,010   \\
        \texttt{electricliterature.com}                  & 2,177    & 28 & 0.29 & 610     & 51 & 0.64 & 1,110   \\
        \texttt{kirkusreviews.com}                       & 68,352   & 93 & 0.99 & 63,567  & 42 & 0.43 & 28,708  \\
        \texttt{publishersweekly.com}                    & 97,337   & 78 & 0.78 & 75,923  & 73 & 0.77 & 71,056  \\
        \texttt{theguardian.com/books}                   & 69,519   & 53 & 0.62 & 36,845  & 38 & 0.45 & 26,417  \\
        \texttt{washingtonpost.com/entertainment/books}  & 1,597    & 66 & 0.76 & 1,054   & 64 & 0.88 & 1,022   \\
        \midrule
        \textbf{Total}                                   & 720,505  &    &      & 301,222 &    &      & 346,558 \\
        \bottomrule
    \end{tabular}%
    }
    \label{tab:book_sources}
\end{table}

\section{Book Corpus Details}
\label{sec:books_used}

Throughout our experiments, we used a set of 100 books, 20 representative books chosen from 5 categories. In \Cref{tab:books_1,tab:books_2} we enumerate the books, their metadata, and the criterion we used to select them for the core book set. 

\begin{table}[htbp]
\caption{Metadata of books used in experiments (1--50), adapted from the Goodreads dataset \citep{goodreads2018recommender}}
\centering
\small
\begin{tabular}{@{}llrrlr@{}}
\toprule
Title & Author & Reviews & Avg Score & Genre & Category \\
\midrule
Cinder & Marissa Meyer & 36235 & 4.11 & fantasy & most reviews \\
Dead Ever After & Charlaine Harris & 7415 & 2.92 & fantasy & lowest average \\
New Moon & Stephenie Meyer & 40992 & 3.26 & young adult & highest std \\
Marked & P.C. Cast & 12092 & 3.14 & young adult & lowest average \\
The Secret History & Donna Tartt & 11596 & 3.80 & mystery & highest std \\
The Lost Symbol & Dan Brown & 21569 & 3.13 & mystery & lowest average \\
Me Before You & Jojo Moyes & 51715 & 4.26 & romance & most reviews \\
The Selection & Kiera Cass & 31715 & 3.76 & romance & most reviews \\
The Historian & Elizabeth Kostova & 11613 & 3.32 & history & lowest average \\
Wuthering Heights & Emily Bronte & 17961 & 3.47 & history & lowest average \\
\midrule
Archer's Voice & Mia Sheridan & 6985 & 4.62 & romance & highest average \\
It Ends with Us & Colleen Hoover & 14236 & 4.52 & romance & highest average \\
Origin & Jennifer L. Armentrout & 6017 & 4.42 & romance & highest average \\
Point of Retreat & Colleen Hoover & 6855 & 4.41 & romance & highest average \\
Hopeless & Colleen Hoover & 16800 & 4.38 & romance & highest average \\
Fifty Shades of Grey & E.L. James & 67203 & 2.73 & romance & lowest average \\
Grey & E.L. James & 8114 & 3.20 & romance & lowest average \\
Fifty Shades Darker & E.L. James & 25251 & 3.42 & romance & lowest average \\
Fifty Shades Freed & E.L. James & 13165 & 3.40 & romance & lowest average \\
The Heir & Kiera Cass & 11690 & 3.48 & romance & lowest average \\
Beautiful Disaster & Jamie McGuire & 21970 & 3.74 & romance & highest std \\
The Notebook & Nicholas Sparks & 14929 & 3.64 & romance & highest std \\
Beautiful Bastard & Christina Lauren & 6659 & 3.54 & romance & highest std \\
Emma & Jane Austen & 8855 & 3.60 & romance & highest std \\
Romeo and Juliet & William Shakespeare & 12510 & 3.51 & romance & highest std \\
Pride and Prejudice & Jane Austen & 35918 & 4.20 & romance & most reviews \\
Everything, Everything & Nicola Yoon & 22363 & 3.98 & romance & most reviews \\
Anna and the French Kiss & Stephanie Perkins & 20077 & 4.18 & romance & most reviews \\
The Nightingale & Kristin Hannah & 33361 & 4.47 & history & highest average \\
Unbroken & Laura Hillenbrand & 38878 & 4.50 & history & highest average \\
To Kill a Mockingbird & Harper Lee & 59827 & 4.43 & history & highest average \\
When Breath Becomes Air & Paul Kalanithi & 10779 & 4.40 & history & highest average \\
The Book Thief & Markus Zusak & 77448 & 4.35 & history & highest average \\
Love in the Time of Cholera & Gabriel Garcia Marquez & 11731 & 3.31 & history & lowest average \\
Wolf Hall & Hilary Mantel & 10254 & 3.48 & history & lowest average \\
Abraham Lincoln: Vampire Hunter & Seth Grahame-Smith & 10031 & 3.54 & history & lowest average \\
Atonement & Ian McEwan & 12609 & 3.61 & history & highest std \\
Outlander & Diana Gabaldon & 30463 & 3.78 & history & highest std \\
A Tale of Two Cities & Charles Dickens & 10362 & 3.86 & history & highest std \\
The Pillars of the Earth & Ken Follett & 23342 & 3.89 & history & highest std \\
The Diary of a Young Girl & Anne Frank & 17771 & 4.11 & history & highest std \\
The Help & Kathryn Stockett & 71318 & 4.40 & history & most reviews \\
All the Light We Cannot See & Anthony Doerr & 49416 & 4.27 & history & most reviews \\
Water for Elephants & Sara Gruen & 51195 & 3.90 & history & most reviews \\
The Guernsey Literary Society & Mary Ann Shaffer & 34373 & 4.14 & history & most reviews \\
11/22/63 & Stephen King & 25838 & 4.18 & history & most reviews \\
Big Little Lies & Liane Moriarty & 27346 & 4.23 & mystery & highest average \\
And Then There Were None & Agatha Christie & 12618 & 4.24 & mystery & highest average \\
Career of Evil & Robert Galbraith & 7743 & 4.11 & mystery & highest average \\
Doctor Sleep & Stephen King & 11885 & 4.11 & mystery & highest average \\
\bottomrule
\end{tabular}
\label{tab:books_1}
\end{table}

\begin{table}[htbp]
\caption{Metadata of books used in experiments (51--100), adapted from the Goodreads dataset \citep{goodreads2018recommender}}
\centering
\small
\begin{tabular}{@{}llrrlr@{}}
\toprule
Title & Author & Reviews & Avg Score & Genre & Category \\
\midrule
NOS4A2 & Joe Hill & 6646 & 4.11 & mystery & highest average \\
Luckiest Girl Alive & Jessica Knoll & 8582 & 3.12 & mystery & lowest average \\
The Dinner & Herman Koch & 11691 & 3.15 & mystery & lowest average \\
The Woman in Cabin 10 & Ruth Ware & 12871 & 3.47 & mystery & lowest average \\
The Da Vinci Code & Dan Brown & 33535 & 3.47 & mystery & lowest average \\
Angels \& Demons & Dan Brown & 19683 & 3.70 & mystery & highest std \\
The Girl with the Dragon Tattoo & Stieg Larsson & 48632 & 3.75 & mystery & highest std \\
The Good Girl & Mary Kubica & 10018 & 3.52 & mystery & highest std \\
Inferno & Dan Brown & 29204 & 3.44 & mystery & highest std \\
The Girl on the Train & Paula Hawkins & 78438 & 3.56 & mystery & most reviews \\
Gone Girl & Gillian Flynn & 69096 & 3.80 & mystery & most reviews \\
The Cuckoo's Calling & Robert Galbraith & 22980 & 3.71 & mystery & most reviews \\
Dark Places & Gillian Flynn & 24433 & 3.78 & mystery & most reviews \\
The Girl Who Played with Fire & Stieg Larsson & 24868 & 4.01 & mystery & most reviews \\
The Hate U Give & Angie Thomas & 10729 & 4.66 & young adult & highest average \\
Crooked Kingdom & Leigh Bardugo & 8674 & 4.58 & young adult & highest average \\
Clockwork Princess & Cassandra Clare & 10423 & 4.57 & young adult & highest average \\
Wonder & R.J. Palacio & 31536 & 4.50 & young adult & highest average \\
Cress & Marissa Meyer & 18007 & 4.50 & young adult & highest average \\
Evermore & Alyson Noel & 8736 & 3.03 & young adult & lowest average \\
Fallen & Lauren Kate & 16254 & 3.06 & young adult & lowest average \\
Crossed & Ally Condie & 7994 & 3.11 & young adult & lowest average \\
Reached & Ally Condie & 9645 & 3.22 & young adult & lowest average \\
Twilight & Stephenie Meyer & 90766 & 3.29 & young adult & highest std \\
Breaking Dawn & Stephenie Meyer & 42587 & 3.38 & young adult & highest std \\
Eclipse & Stephenie Meyer & 32909 & 3.54 & young adult & highest std \\
Midnight Sun & Stephenie Meyer & 9247 & 3.96 & young adult & highest std \\
The Fault in Our Stars & John Green & 129572 & 4.37 & young adult & most reviews \\
The Hunger Games & Suzanne Collins & 142645 & 4.33 & young adult & most reviews \\
Mockingjay & Suzanne Collins & 86946 & 3.82 & young adult & most reviews \\
Catching Fire & Suzanne Collins & 80495 & 4.26 & young adult & most reviews \\
Divergent & Veronica Roth & 68482 & 3.96 & young adult & most reviews \\
A Court of Mist and Fury & Sarah J. Maas & 18765 & 4.63 & fantasy & highest average \\
A Monster Calls & Patrick Ness & 18601 & 4.55 & fantasy & highest average \\
Harry Potter and the Deathly Hallows & J.K. Rowling & 45748 & 4.54 & fantasy & highest average \\
Harry Potter and the Goblet of Fire & J.K. Rowling & 25258 & 4.50 & fantasy & highest average \\
Clockwork Princess & Cassandra Clare & 10861 & 4.55 & fantasy & highest average \\
Pride and Prejudice and Zombies & Seth Grahame-Smith & 11216 & 2.88 & fantasy & lowest average \\
Wicked & Gregory Maguire & 20367 & 2.92 & fantasy & lowest average \\
The Magicians & Lev Grossman & 16275 & 3.00 & fantasy & lowest average \\
The Short Second Life of Bree Tanner & Stephenie Meyer & 8939 & 3.29 & fantasy & lowest average \\
Jonathan Strange \& Mr Norrell & Susanna Clarke & 8374 & 3.57 & fantasy & highest std \\
Hush, Hush & Becca Fitzpatrick & 20227 & 3.60 & fantasy & highest std \\
Eragon & Christopher Paolini & 16700 & 3.39 & fantasy & highest std \\
A Discovery of Witches & Deborah Harkness & 22410 & 3.53 & fantasy & highest std \\
Shiver & Maggie Stiefvater & 18330 & 3.47 & fantasy & highest std \\
The Martian & Andy Weir & 49988 & 4.25 & fantasy & most reviews \\
Miss Peregrine's Home for Peculiar Children & Ransom Riggs & 46665 & 3.66 & fantasy & most reviews \\
Harry Potter and the Cursed Child & John Tiffany & 36121 & 3.47 & fantasy & most reviews \\
City of Bones & Cassandra Clare & 43280 & 3.67 & fantasy & most reviews \\
\bottomrule
\end{tabular}
\label{tab:books_2}
\end{table}

\end{document}